\documentclass[lettersize,journal]{IEEEtran}

\usepackage{times}
\usepackage{epsfig}
\usepackage{graphicx}
\usepackage{amsmath}
\usepackage{amssymb}
\usepackage{color}
\usepackage{subfig}
\usepackage{tabularx}
\usepackage{multirow}
\usepackage[abs]{overpic}
\usepackage{algorithm}
\usepackage{algorithmic}
\usepackage{wrapfig}
\usepackage{ragged2e}
\usepackage[nocompress]{cite}
\usepackage[colorlinks,bookmarksnumbered,bookmarksopen,linkcolor=black,citecolor=black,urlcolor=black]{hyperref}
\usepackage{enumitem} 
\usepackage{multirow} 
 \usepackage{booktabs}

\usepackage[normalem]{ulem}  

\usepackage{geometry}
\usepackage{bbding}

\graphicspath{{./Imgs/}{./CVPR/Imgs/}}
\DeclareGraphicsExtensions{.pdf,.png,.jpg}
\newcolumntype{C}[1]{>{\centering\arraybackslash}p{#1}}

\newcommand{\figref}[1]{Fig.~\ref{#1}}

\def\ie{\emph{i.e.~}}
\def\eg{\emph{e.g.~}}

\begin{document}

\title{Heatmap Pooling Network for Action Recognition from RGB Videos}
\author{Mengyuan Liu, Jinfu Liu, Yongkang Jiang, Bin He 
\thanks{This work was supported by National Key Research and Development Program of China (No. 2024YFB4709800), Guangdong S$\&$T Program (No. 2024B0101050002), Shenzhen Innovation in Science and Technology Foundation for The Excellent Youth Scholars (No. RCYX20231211090248064). (Corresponding Author: Jinfu Liu)}
\thanks{Mengyuan Liu is with the State Key Laboratory of General Artificial Intelligence, Peking University, Shenzhen Graduate School, China (E-mail: liumengyuan@pku.edu.cn). Jinfu Liu is with the Imaging Department, DJI Technology Co., Ltd, China (E-mail: goch.liu@dji.com). Yongkang Jiang and Bin He are with TongJi University, China (E-mail: jiangyongkang@tongji.edu.cn, hebin@tongji.edu.cn).}}

\markboth{IEEE TRANSACTIONS ON PATTERN ANALYSIS AND MACHINE INTELLIGENCE, 2025}%
{Shell \MakeLowercase{\textit{et al.}}: A Sample Article Using IEEEtran.cls for IEEE Journals}


\maketitle

\begin{abstract}
Human action recognition (HAR) in videos has garnered widespread attention due to the rich information in RGB videos. Nevertheless, existing methods for extracting deep features from RGB videos face challenges such as information redundancy, susceptibility to noise and high storage costs. To address these issues and fully harness the useful information in videos, we propose a novel heatmap pooling network (HP-Net) for action recognition from videos, which extracts information-rich, robust and concise pooled features of the human body in videos through a feedback pooling module. The extracted pooled features demonstrate obvious performance advantages over the previously obtained pose data and heatmap features from videos. In addition, we design a spatial-motion co-learning module and a text refinement modulation module to integrate the extracted pooled features with other multimodal data, enabling more robust action recognition. Extensive experiments on several benchmarks namely NTU RGB+D 60, NTU RGB+D 120, Toyota-Smarthome and UAV-Human consistently verify the effectiveness of our HP-Net, which outperforms the existing human action recognition methods. Our code is publicly available
at: \href{https://github.com/liujf69/HPNet-Action}{https://github.com/liujf69/HPNet-Action}.
\end{abstract}


\maketitle

\section{Introduction}\label{sec:introduction}
\IEEEPARstart{H}{uman} action recognition (HAR) \cite{Feichtenhofer_2019_ICCV, LIU2017346, PoseC3D, MMNet} is an active research task in video understanding, as human actions convey essential information like body movements and tendencies, providing valuable insights into understanding individuals in videos \cite{liu2024mmcl, reilly2024pivit, song2022constructing, cheng2024dense}. The HAR also holds significant research value and has broad applications in fields like virtual reality, human-robot interaction, and content-based video retrieval. To achieve accurate action recognition, various video modalities have been extensively studied, including RGB frames \cite{PoseC3D, MMNet}, depth images \cite{9422825}, pose sequences \cite{yan2018spatial, Chen_2021_ICCV, 10113233}, textual descriptions \cite{ActionClip, xu2023language}, human parsing \cite{EPPNet} and point cloud \cite{deng2024vg4d}. Specifically, RGB video frames contain abundant information, encompassing body actions, physical appearances and objects interacting with individuals, making them a highly favored modality for action recognition. Consequently, a series of classic action recognition methods \cite{carreira2017quo, Feichtenhofer_2019_ICCV, PoseC3D} based on RGB videos have been developed, leveraging this rich visual information to enhance recognition accuracy and robustness.

\begin{figure}[t]
  \centering
   \includegraphics[width=1.0\linewidth]{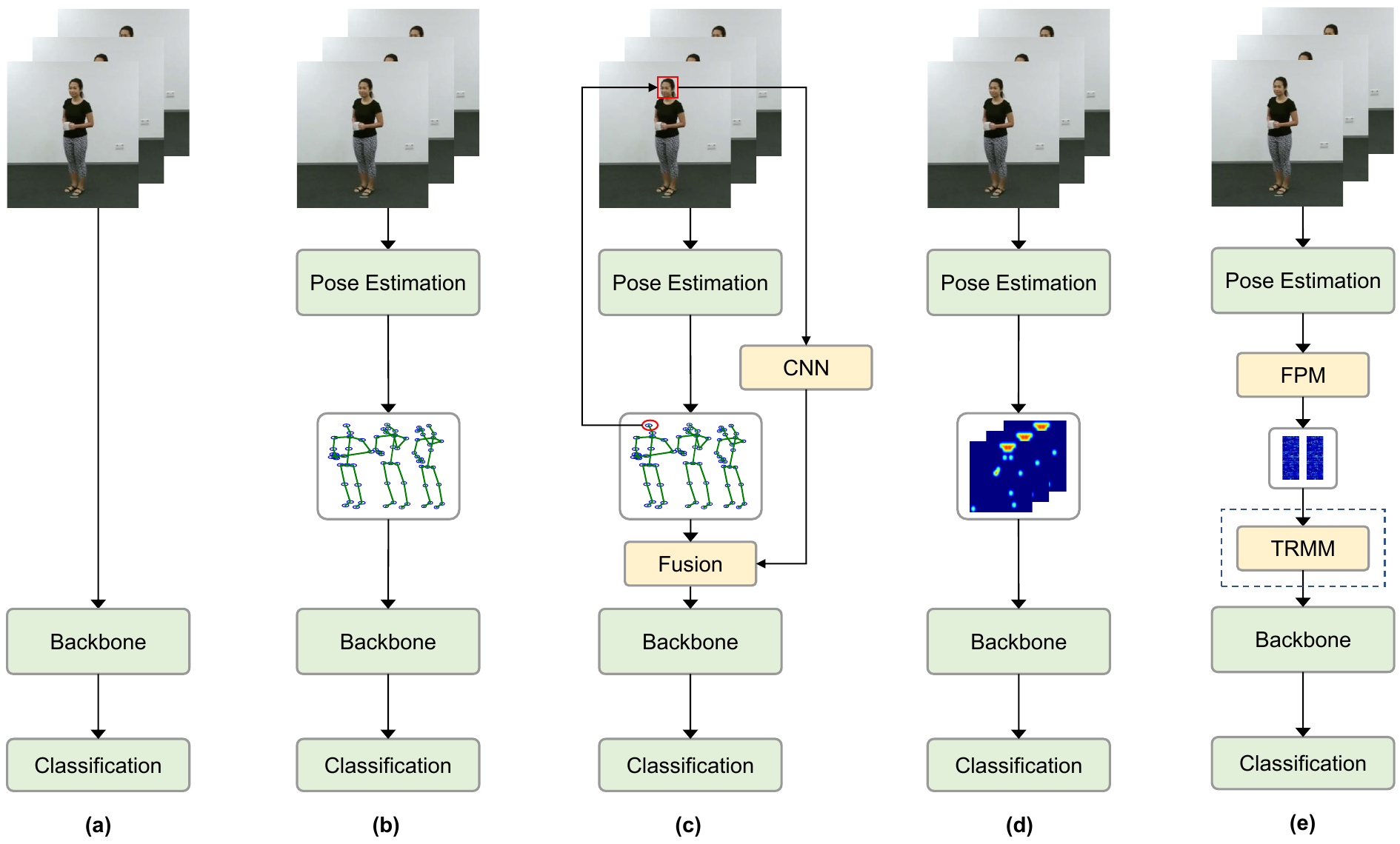}
   \vspace{-2em}
   \caption{\textbf{Comparison between existing methods and ours.} Existing methods for modeling features from RGB videos have several limitations. (a) RGB-based methods rely on RGB images that are prone to environmental interference and noise. (b) Pose-based methods use pose data with limited information, making it difficult to support fine-grained action recognition. (c) Pose-guided feature aggregation methods require additional modeling and fusion networks, resulting in higher computational cost. (d) Heatmap-based methods often involve redundant data and incur high storage costs. (e) Our proposed HP-Net incorporates a Feedback Pooling Module (FPM) to obtain more efficient heatmap representations by reducing redundancy while retaining relevant information. In addition, the Text Refinement Modulation Module (TRMM) is used to refine the visual features with textual guidance, helping to further improve representation quality.}
   \vspace{-1em}
   \label{fig:figure1}
\end{figure}

Existing methods \cite{yan2018spatial, Feichtenhofer_2019_ICCV, PoseC3D, Liu_2018_CVPR} for modeling features from RGB videos can be broadly classified into four main categories. A straightforward approach involves using different visual backbones (\eg CNNs and ViTs) to directly model video frames \cite{Feichtenhofer_2019_ICCV, XCLIP, wang2022internvideo}, as illustrated in \figref{fig:figure1} (a), ultimately relying on the extracted deep features for action recognition. Although these RGB-based methods that directly utilize raw video frames as input can almost use all the information in videos, they are susceptible to environmental interference and other intrinsic noise in images, which may hinder recognition performance. The second common approach involves using pose estimation to extract 2D/3D pose data from videos and performing action recognition based on the estimated poses \cite{yan2018spatial, reilly2024pivit, kim2023cross}, as illustrated in \figref{fig:figure1} (b). Although these 2D/3D pose-based methods can significantly alleviate environmental interference \cite{yan2018spatial, Chen_2021_ICCV, 10113233, xiang2023gap, zhou2023learning}, they omit a wealth of valuable information (\eg appearance and objects) and are challenging to achieve fine-grained action recognition. The third common approach leverages pose information to guide the extraction of RGB patches, which are modeled using CNNs and then fused with pose features for human action recognition, as illustrated in \figref{fig:figure1} (c). However, these pose-guided feature aggregation methods require additional RGB patch modeling and multimodal fusion modules, often resulting in higher computational costs.

To address the issues present in the above three approaches, the fourth effective approach is to retain the heatmap features obtained from human pose estimation \cite{wang2020deep, li2022simcc, li2024hourglass} in videos, as illustrated in \figref{fig:figure1} (d), and further model them using different visual backbones \cite{PoseC3D, Liu_2018_CVPR} for action recognition. These heatmap-based methods \cite{PoseC3D, Liu_2018_CVPR} combine the advantages of the three approaches mentioned above and have been proven to be an effective solution for action recognition. Nevertheless, existing heatmap-based methods typically store heatmaps in image format and model them using image-based visual backbones, which still face challenges including information redundancy, high storage costs, and limitations imposed by downstream modeling backbones tailored to specific input formats. The aforementioned heatmap-based methods naturally lead to a question: \textit{How to better model heatmaps to obtain more concise features, reduce redundancy and support a wider range of visual backbones?} Motivated by the above question, we propose a novel heatmap pooling network (HP-Net) for action recognition from videos, which extracts information-rich, robust and concise heatmap pooled features from videos via a feedback pooling module (FPM), as illustrated in \figref{fig:figure1} (e). 

To better and more intuitively explain the motivation of our work, we begin by reviewing the classical heatmap-based pose estimation methods \cite{wang2020deep, li2022simcc}. These methods use neural networks to model raw video frames in a layer-by-layer manner, which inevitably leads to both information compression and information loss. Throughout this process, heatmaps of various stages and scales are generated, which retain the spatial structure of the human body and represent the probability distributions of human poses using 2D Gaussian functions. To effectively capture human motion in videos, heatmap-based pose estimation methods typically output simplified human pose coordinates, which are then used for various downstream tasks in human motion understanding, such as human action recognition in this paper. In detailed implementation, these pose estimation methods project multi-scale heatmaps into a fixed number of channels through convolutional operations, with each channel corresponding to a specific pose to be estimated. The values within each channel represent the likelihood of the pose occurring at each position, while the final pose locations are then determined using an Argmax operation to select the coordinates with the highest probability. Although these pose estimation methods produce concise human poses to describe human motion in videos, they suffer from three critical information loss issues for action recognition tasks: (1) Feature Reduction: The final human pose coordinates discard appearance features (\eg color and texture) that could enhance motion understanding, retaining only sparse positional data. (2) Channel Compression: heatmaps with diverse scales and channel counts are forcibly projected to a fixed channel number (\eg 17 for the COCO dataset), which may lead to the loss of many valuable channel-wise features. (3) Spatial Oversimplification: The Argmax operation retains only the highest-probability coordinate per heatmap, ignoring neighboring regions that may contain valid cues for human motion.

To mitigate information loss, prior heatmap-based action recognition methods \cite{Liu_2018_CVPR, PoseC3D} directly feed entire heatmaps into downstream visual networks, introducing severe information redundancy, as most heatmap pixels represent irrelevant or noisy coordinate probabilities. In contrast, our proposed heatmap pooling network (HP-Net) rethinks the core principles of heatmap-based pose estimation. By introducing a feedback pooling mechanism to generate heatmap pooled features, our HP-Net strategically aggregates features from multi-stage and multi-scale heatmaps guided by estimated pose positions. Intuitively, the estimated poses inherently localize the most informative regions in heatmaps. Centering pooling operations around these poses maximizes the retention of discriminative local features, while preserving all channel dimensions across heatmaps in different scales. This design in our HP-Net offers two key advantages: (1) Targeted Feature Preservation: Human pose positions, being probability maxima, anchor regions rich in motion-relevant features. Pooling around these pose points focuses on meaningful areas, suppressing noise while retaining critical spatial-contextual details. (2) Multi-Channel Integrity: By avoiding channel compression and leveraging full-channel heatmaps, the HP-Net captures latent features across hierarchical representations, essential for modeling complex human motions. Notably, our HP-Net utilizes a feedback pooling mechanism, where the estimated concise pose coordinates are used to reversely pool the information-redundant heatmaps from different stages and scales, enabling the extraction of full-channel and information-rich heatmap-pooled features. Thus, the HP-Net fundamentally reconciles the trade-off between information loss and redundancy, enabling efficient yet comprehensive motion representation for human motion analysis. 

Compared to conventional pose data and image-based heatmap features extracted from videos, our heatmap pooled features offer more useful information, less redundancy and lower storage costs, while being compatible with a wider range of visual backbones, such as graph convolutional networks (GCNs) and Transformers. Although modeling heatmap pooled features through visual networks outperforms approaches that directly process raw heatmaps or pose data, it still falls short of current state-of-the-art (SOTA) multimodal action recognition methods \cite{reilly2024pivit, xu2023language, PoseC3D, MMNet}. In essence, the heatmap-pooled features obtained through the feedback pooling mechanism are still unimodal in nature and thus inherently subject to the limitations of single-modality representations. Therefore, the recognition performance only based on the heatmap pooled features is inevitably inferior to that of existing multimodal approaches. We also observe that these leading multimodal methods \cite{reilly2024pivit, xu2023language, PoseC3D, MMNet} typically design specialized modules to fuse multimodal data (\eg human pose, RGB video frames or text descriptions), leveraging cross-modality complementarity to enhance recognition accuracy. To validate the transferability of our heatmap-pooled features and demonstrate their potential to complement and integrate with other modalities, our proposed HP-Net futher integrates the extracted pooled features with other modalities, such as deep features from RGB video frames and text features generated by large language models (LLMs), enabling more robust and accurate action recognition. In the detailed implementation, we propose two modules to enable feature-level multimodal fusion of heatmap pooled features, RGB features and textual features, while also exploring the effectiveness of decision-level fusion through a multi-stream ensemble strategy.

In our heatmap pooling network (HP-Net), we employ a feedback pooling module (FPM) to perform pooling operation on heatmap features using 2D pose data obtained from pose estimation. These pooled features can be effectively modeled by using backbones that accept topology graph inputs (\eg GCNs), thereby avoiding the use of complex visual backbones designed exclusively for heatmap images (\eg 3D CNNs). Meanwhile, we designed a text refinement modulation module (TRMM) to adapt the pooled features for refining and modulating text features. These refined and modulated text features are then further fused with video features, demonstrating that heatmap pooled features can be effectively integrated with multimodal features to achieve more robust action recognition. In addition, we designed a spatial-motion co-learning module (SMCLM) to transform heatmap pooled features into spatiotemporal modalities, which are modeled by using different graph convolutional networks (GCNs), further enhancing the recognition performance. 

\begin{figure*}[t]
  \centering
   \includegraphics[width=1.0\linewidth]{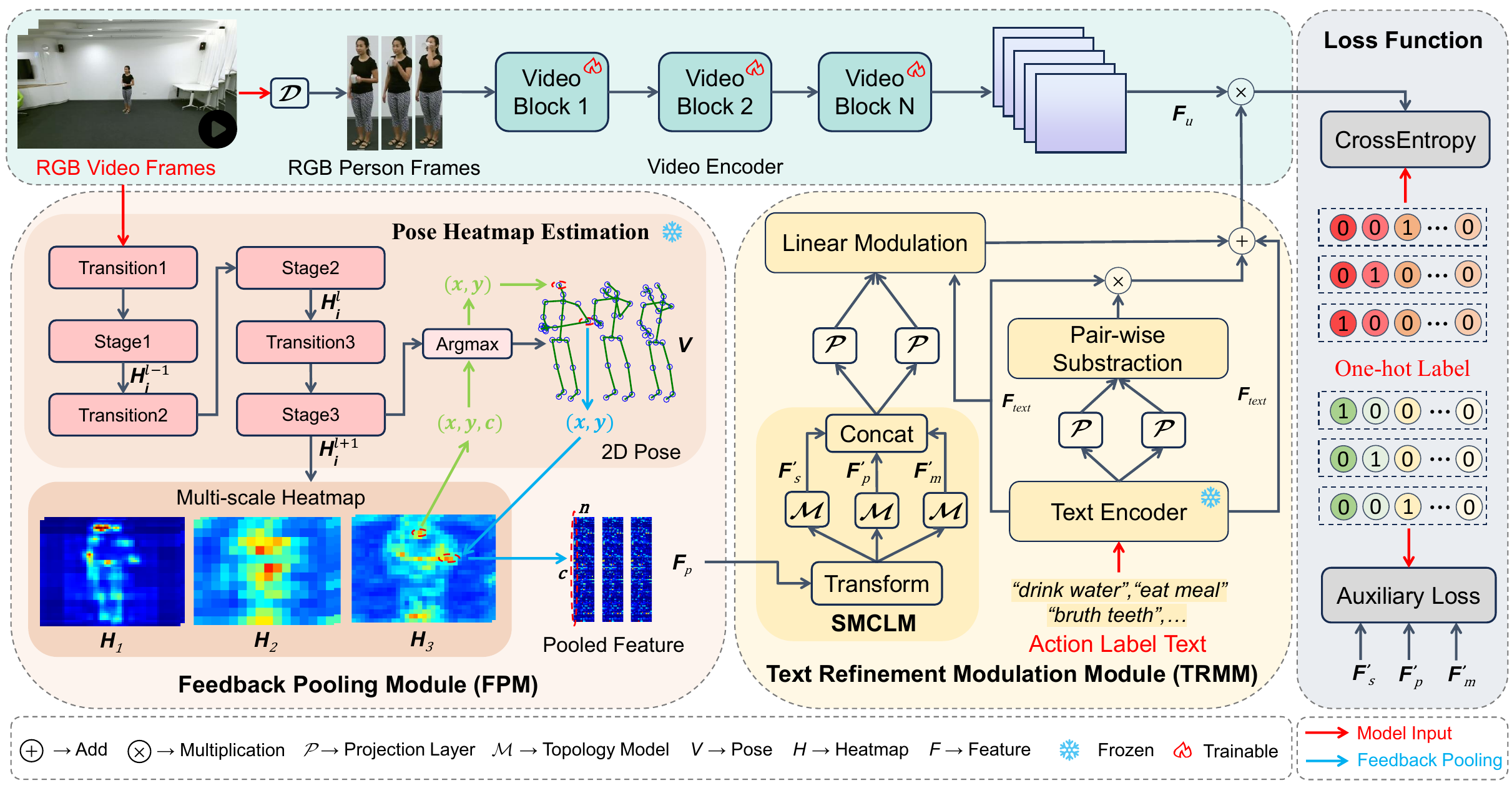}
   \caption{\textbf{Framework of our proposed Heatmap Pooling Network (HP-Net).} We utilize a feedback pooling module (FPM) to extract information-rich, robust and concise heatmap pooled features from videos. The spatial-motion co-learning module (SMCLM) is designed to model spatial and motion dynamics within heatmap pooled features, enabling the generation of spatiotemporal representations that seamlessly integrate with other modalities. The text refinement modulation module (TRMM) aims to enhance and modulate the basic text features of action labels, transforming them into enriched semantic representations that effectively support human action recognition.}
   \label{fig:figure2}
\end{figure*}

The framework of our heatmap pooling network (HP-Net) is shown in \figref{fig:figure2} and our contributions are summarized as follows:

\setlist{nolistsep}
\begin{itemize}[noitemsep, leftmargin=*]
\item We propose a novel heatmap pooling network (HP-Net) for action recognition from videos, which extracts concise, robust and transferable heatmap features through a feedback pooling mechanism. The proposed feedback pooling module (FPM) in our HP-Net is orthogonal to the heatmap-based human pose estimation methods, providing a new connection bridge between the fields of human pose estimation and action recognition.
\item Our HP-Net further enhances image-based human heatmap features, making them adaptable to a wider range of visual backbone networks. Simultaneously, the extracted heatmap pooled features can be smoothly integrated with other multimodal features, facilitating more accurate and robust human action recognition. Compared to traditionally acquired 2D pose data, the proposed heatmap pooled feature achieves consistent performance gains when integrated with distinct GCN/Transformer architectures. 
\item Our HP-Net conduct the most extensive evaluation of heatmap-based human action recognition, spanning three major benchmarks with radically different characteristics: controlled laboratory settings (\ie NTU-60 and NTU-120), home environments (\ie Toyota-Smarthome) and aerial surveillance (\ie UAV-Human). The proposed HP-Net achieves state-of-the-art performance in all the above scenarios, providing compelling evidence of its effectiveness and generalizability. 
\end{itemize}

The remainder of this paper is organized as follows. Section 2 briefly reviews related work. Section 3 describes the structure of the heatmap pooling network. Section 4 reports the experimental results and analysis. Section 5 discusses the generalizability of our work. Section 6 concludes the paper.

\section{Related Work}
The task of human action recognition from videos has attracted significant attention from researchers due to the availability of rich multimodal data. As a result, a series of mainstream methods have emerged based on different modalities. In this section, we will first discuss the related works in both RGB-based methods and pose-based methods. Following that, we will introduce heatmap-based methods, which are more relevant to the work of this paper. Finally, we will provide an overview of the classic upstream task in human action recognition, namely human pose estimation from RGB videos.

\subsection{RGB-based Human Action Recognition}
RGB-based human action recognition have garnered widespread attention due to the RGB videos contain rich information and are relatively simple to obtain in daily life \cite{liu2017pku, wang2014cross, shao2020finegym, soomro2012ucf101, kuehne2011hmdb}. Due to the unique advantages of convolutional neural networks (CNNs) in processing Euclidean data, most previous works \cite{carreira2017quo, Feichtenhofer_2019_ICCV, das2020vpn, MMNet} have used CNNs to model the spatiotemporal features of RGB videos. For instance, Carreira et al. \cite{carreira2017quo} introduced the inflated 3D CNN (I3D) that leveraging both RGB and optical flow streams for human action recognition. Feichtenhofer et al. \cite{Feichtenhofer_2019_ICCV} presented the SlowFast networks for video action recognition, which involves a slow pathway to capture spatial semantics and a fast pathway to capture motion of RGB videos. Additionally, some works \cite{joze2020mmtm, bruce2021multimodal, das2020vpn, MMNet} have fused features from RGB video frames with pose features to achieve more robust human action recognition. Das et al. \cite{das2020vpn} proposed the VPN, which models spatiotemporal features of RGB video frames using a visual backbone and integrates RGB features with pose features through spatial embedding and an attention network. Yu et al. \cite{MMNet} transformed RGB video into a single-image ST-ROI and proposed the MM-Net, which uses pretrained CNNs to model ST-ROI and fuses it with pose features for video action recognition. Reilly et al. \cite{reilly2024pivit} introduce the first Pose Induced Video Transformer ($\pi$ ViT), which augments the RGB representations learned by video transformers with pose information for action recognition. The aforementioned methods directly use RGB video frames as input to visual networks, making them susceptible to the inherent noise in RGB frames, which reduces the accuracy of human action recognition.

\subsection{Pose-based Human Action Recognition}
Human skeletons/poses have become a commonly used modality in action recognition due to their simplicity and resistance to environmental interference \cite{10694798, yang2024one, yang2023self}. Generally, human poses can be obtained from videos through sensors \cite{7780484, 8713892, kay2017kinetics} or pose estimation \cite{wang2020deep, li2022simcc, li2024hourglass}. These human poses extracted from videos is a structured natural topology graph, making them well-suited for modeling with graph convolutional networks (GCNs) in pose-based action recognition. For instance, Yan et al. \cite{yan2018spatial} were the first to introduce GCNs for pose-based action recognition, proposing the ST-GCN to capture spatiotemporal features from pose data. Shi et al. \cite{shi2019two} proposed a novel two-stream adaptive graph convolutional network (2s-AGCN), which model both the first-order and the second-order information of human pose simultaneously. Chen et al. \cite{Chen_2021_ICCV} enhanced the design of GCNs by introducing the channel-wise topology refinement graph convolutional network (CTR-GCN), which efficiently aggregates joint features within each channel. Liu et al. \cite{10113233} proposed the temporal decoupling graph convolutional network (TD-GCN), which innovatively utilizes a temporal-dependent adjacency matrix to effectively extract high-level features from pose data. Since human pose is merely a set of simple coordinate data with limited information, the aforementioned pose-based methods have drawbacks in fine-grained action recognition. 

\begin{figure*}[t]
  \centering
   \includegraphics[width=1.0\linewidth]{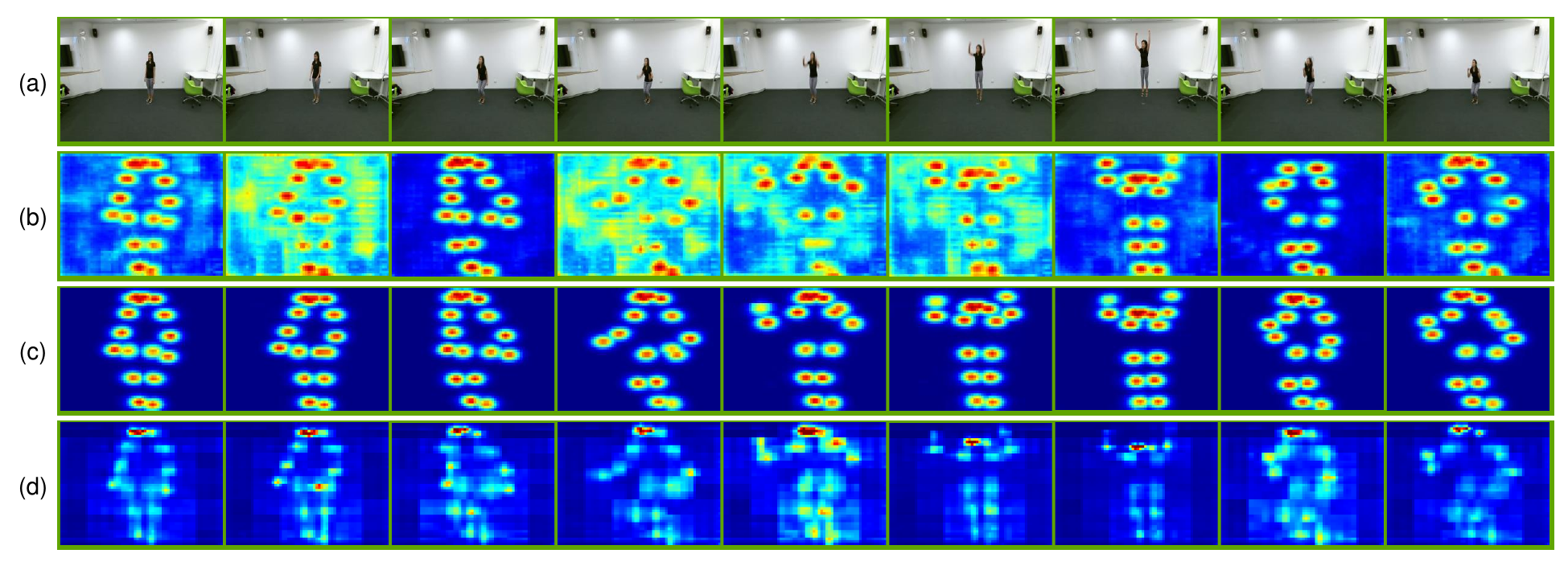}
   \caption{Visualization of human heatmaps obtained by different human pose estimation models. (a) Visualization of the video frame sequence. (b) Visualization of human heatmaps obtained using ResNet \cite{xiao2018simple}. (c) Visualization of human heatmaps obtained using HR-Net \cite{wang2020deep}. (d) Visualization of human heatmaps obtained using SimCC \cite{li2022simcc}.}
   \label{fig:multi_frames}
\end{figure*}

\subsection{Heatmap-based Human Action Recognition}
Human heatmap as a feature representation in the human pose estimation process \cite{wang2020deep, li2022simcc}, can filter environmental interference while containing most information about human action. Some previous works \cite{Liu_2018_CVPR, PoseC3D} have used heatmaps for video action recognition and achieved significant success. For instance, to eliminate the limitations of cluttered backgrounds and non-action motions for video action recognition, Liu et al. \cite{Liu_2018_CVPR} were the first to introduce heatmaps as a human cue for action recognition. To address the sparse nature of heatmaps obtained from the pose estimation process, Liu et al. \cite{Liu_2018_CVPR} also developed a spatial rank pooling method to aggregate the evolution of heatmaps into body shape evolution images. In addition, Duan et al. \cite{PoseC3D} proposed the PoseConv3D, which relies on 3D heatmap volumes as input and uses the pretrained 3D CNNs to effectively model spatiotemporal features. The above methods typically convert heatmaps into image formats as model inputs to facilitate effective modeling with CNNs, which to some extent results in model limitations and information redundancy. Unlike the aforementioned methods that convert heatmaps into image formats, our HP-Net employs feedback pooling to transform redundant heatmaps into topological pooled features, enabling efficient modeling with compact networks that accept topological inputs.

\subsection{Human Pose Estimation from RGB Videos}
Human pose estimation (HPE) \cite{cao2017realtime, fang2022alphapose} aims to localize body joints from a single RGB image and has garnered significant attention as an upstream task for various vision applications. The pose data obtained from videos through human pose estimation can be used for pose-based action recognition. Additionally, heatmap-based methods \cite{wang2020deep, li2022simcc} is a key branch of human pose estimation, which produce heatmaps that can be effectively used for video action recognition \cite{Liu_2018_CVPR, PoseC3D}. Some previous works have explored how to effectively estimate human pose from videos based on heatmaps. For instance, Sun et al. \cite{wang2020deep} proposed the HRNet, which learns reliable high-resolution representations and integrates features of different scales to predict accurate keypoint heatmaps. Li et al. \cite{li2022simcc} proposed the SimCC, which reinterprets the traditional 2D heatmap-based pose estimation approach as two classification tasks for horizontal and vertical coordinates, achieving better prediction performance. The heatmaps generated by the aforementioned pose estimation methods effectively eliminate environmental interference while preserving critical cues about human motion. Consequently, these heatmaps can be readily applied to the human action recognition and seamlessly integrated with other modalities to enhance recognition performance. In \figref{fig:multi_frames}, we visualize the human heatmaps for one action sample using different pose estimation models \cite{wang2020deep, li2022simcc, xiao2018simple}. In our proposed heatmap pooling network (HP-Net), we employ a feedback pooling mechanism to transform heatmaps obtained from pose estimation into concise and robust pooled features. By effectively integrating these pooled features with RGB video and text features, our HP-Net achieve more accurate video action recognition.

\begin{figure*}[t]
  \centering
   \includegraphics[width=1.0\linewidth]{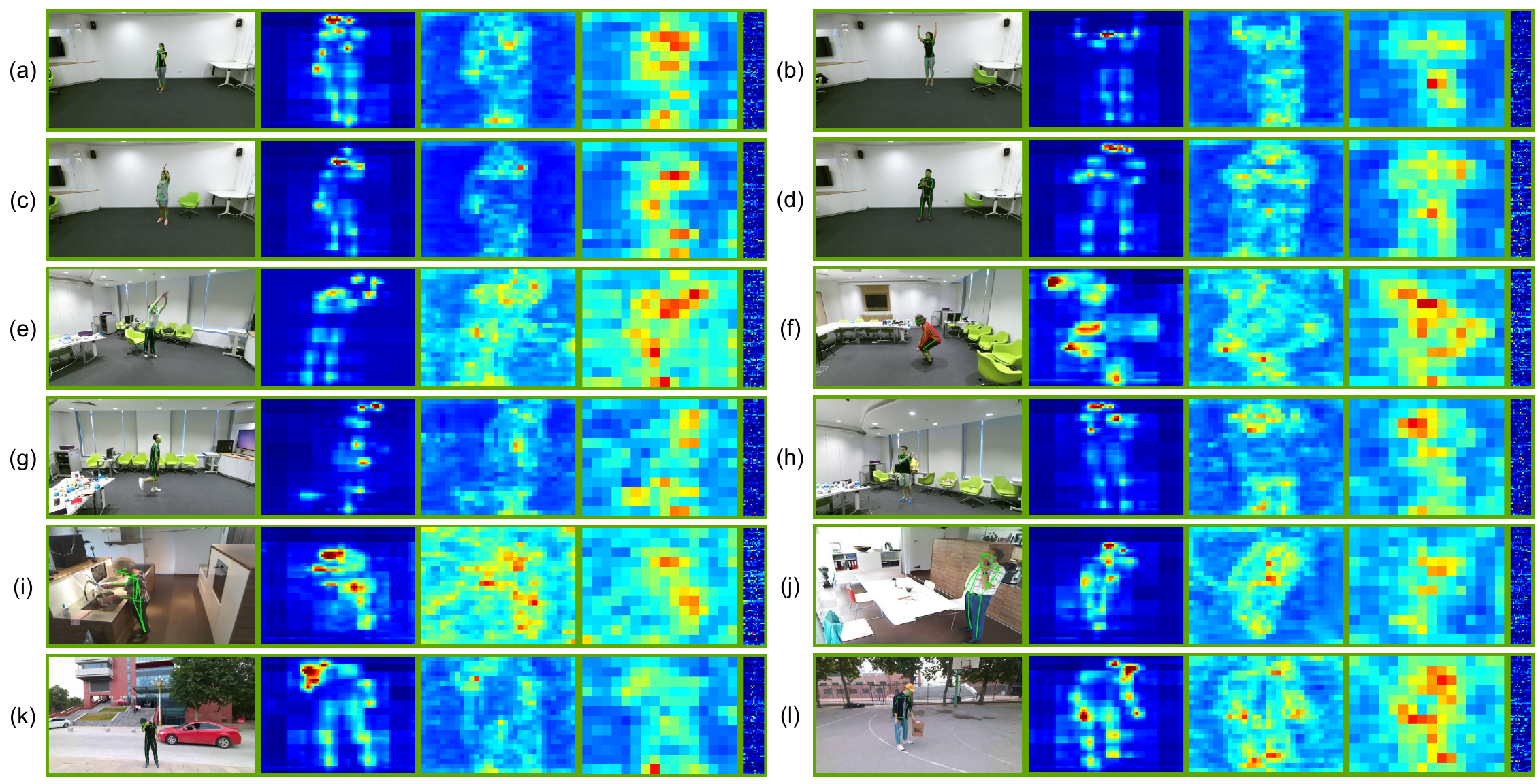}
   \caption{Visualization of different action samples. In each subplot, the first column represents the visualization of the 2D pose data obtained from human pose estimation on the original RGB video frame. The second, third and fourth columns correspond to the human heatmaps $H_1$, $H_2$ and $H_3$, respectively. The fifth column shows the heatmap pooled features output by our Feedback Pooling Module (FPM).}
   \label{fig:heatmap}
\end{figure*}

\section{Heatmap Pooling Network}
In this section, we present our proposed Heatmap Pooling Network (HP-Net) in detail. HP-Net aims to extracts information-rich, robust and concise heatmap pooled features in videos for human action recognition. Meanwhile, the extracted heatmap pooled features are adaptable to different network backbones and can be effortlessly integrated with other modalities, enabling more precise recognition performance. Below we will introduce our HP-Net in five parts. In the first part, we will introduce the detailed implementation of effectively extracting heatmap pooled features from videos using the Feedback Pooling Module (FPM). Then in the second part, we introduce the method of using lightweight networks for topological modeling of heatmap pooled features. Additionally, we provide a detailed explanation of how heatmap pooled features are transferred and integrated with other modalities through the SMCLM and TRMM in the third and fourth parts. Finally, in the fifth section, we introduce the method for obtaining robust classification scores through multimodal fusion and define the loss function of our proposed HP-Net.

\subsection{Feedback Pooling Module}
In our proposed Heatmap Pooling Network (HP-Net), we employ a Feedback Pooling Module (FPM) to extract heatmap pooled features from videos. The FPM aims to leverage the compact pose data obtained from pose estimation to acquire robust and information-rich human heatmap pooled features by reflectively pooling the human heatmap containing human motion cues. In terms of implementation details, we use a pose estimation model $\phi$ to obtain multi-scale human heatmaps based on the following:
\begin{equation}
\begin{split}
    \textbf{\textit{H}}^t &= \{\textbf{\textit{H}}^t_{s_1}, \textbf{\textit{H}}^t_{s_2}, \dots, \textbf{\textit{H}}^t_{s_n}\} \\ &= \phi(\textbf{\textit{U}}^t, \{s_1, s_2, \dots, s_n\}), \quad \forall t \in \{1, 2, \dots, T\},
\end{split}
\label{eq:HPE}
\end{equation}
where $\textbf{\textit{H}}^t$ and $\textbf{\textit{U}}^t$ represent the obtained human heatmap and the original video frame at frame $t$, respectively. The $\{s_1, s_2, \dots, s_n\}$ represent parameter sets of different scales. Then, as is common in most heatmap-based pose estimation methods \cite{wang2020deep, li2022simcc, qu2022heatmap}, we apply the $argmax$ operation along the channel dimension of the heatmap $\textbf{\textit{H}} \in \mathbb{R}^{h \times w \times c}$ to obtain the concise human pose coordinates, which is formulated as:
\begin{equation}
    \textbf{\textit{V}}_i = \arg\max_{(x, y)} \textbf{\textit{H}}_i(x, y), \quad \forall i \in \{1, 2, \dots, n\},
    \label{eq:pose}
\end{equation}
where $\textbf{\textit{V}} = \{(x_1, y_1), (x_2, y_2), \dots, (x_c, y_c)\}$ represents the 2D coordinate set of the human pose. The $h$, $w$ and $c$ represent the height, width and channels of the heatmap $\textbf{\textit{H}} \in \mathbb{R}^{h \times w \times c}$ respectively. Finally, we perform a feedback pooling operation on the human heatmaps of different scales using the extracted 2D pose to obtain concise heatmap pooled features. The feedback pooling operation $\mathcal{G}$ is formulated as:
\begin{equation}
    \textbf{\textit{F}}_{\text{p}}(x, y) = \mathcal{G}\left( \textbf{\textit{H}}_{(x - \frac{R}{2}): (x + \frac{R}{2}), (y - \frac{R}{2}): (y +      \frac{R}{2})} \right),
    \label{eq:FPM}
\end{equation}
where $\textbf{\textit{F}}_{\text{p}}(x, y)$ represents the heatmap pooled features corresponding to the pose coordinates $(x, y)$ and $R$ represents the size of the pooled region. Typically, we pool the human heatmap using $n$ key pose coordinates, resulting in heatmap pooled features $\textbf{\textit{F}}_{\text{p}} \in \mathbb{R}^{n \times c}$ corresponding to $n$ human joints. Based on the above formulas, our proposed FPM can extract compact heatmap pooled features $\textbf{\textit{F}}_{\text{p}}$ related to human action from videos. The extracted heatmap pooled features maintain the simplicity of human pose while significantly reducing the information redundancy in the human heatmaps, preserving rich and effective features related to human action. It is worth emphasizing that in the proposed FPM, the final obtained pose data can be used to reverse-pool heatmaps from different stages and scales, enabling the extraction of diverse human motion cues.

\subsection{Topology Modeling}
The heatmap pooled features extracted by the FPM have the same topological structure as human pose, allowing for effective modeling using classical topological models like GCNs, while avoiding the use of heavy CNNs. Meanwhile, many previous works \cite{wang2020deep, li2022simcc, liu2024mmcl} have used GCNs to model the human poses with topological structure, achieving great success in pose-based action recognition. Therefore, to ensure a fair comparison with human poses and to demonstrate the superiority of the heatmap pooled features, our HP-Net employs a lightweight GCN to perform topological modeling of the extracted heatmap pooled features. Here we select GCN as the topological model $\mathcal{M}$ and the normal graph convolution of GCN can be formulated as:
\begin{equation}
  \textbf{\textit{F}}^{l+1} = \sigma\left(\textbf{\textit{D}}^{-\frac{1}{2}}\textbf{\textit{A}}\textbf{\textit{D}}^{-\frac{1}{2}}\textbf{\textit{F}}^{l}\textbf{\textit{W}}^{l} \right),
  \label{eq:gcn}
\end{equation}
where $\textbf{\textit{F}}^{l}$ is the heatmap pooled features at layer $\textit{l}$ and $\sigma$ is the activation function. $\textbf{\textit{A}}$ is the adjacency matrix representing topological connections. $\textbf{\textit{D}}$ is the degree matrix and $\textbf{\textit{W}}^{\textit{l}}$ is the learnable parameter of the $\textit{l}$-th layer. In our HP-Net, the topologically modeled heatmap pooled features can be directly used for action classification through a simple classification head or transferred for fusion with other modalities to achieve multimodal-based action recognition. It is worth emphasizing that the heatmap pooled features we propose are orthogonal to the backbone networks and thus can be coupled with various topological models.

\subsection{Spatial-motion Co-learning Module}
Human action recognition is a task with strong spatiotemporal correlations and it is a helpful approach to addressing this task effectively via generating spatiotemporal features. In our HP-Net, we utilize a Spatial-motion Co-learning Module (SMCLM) to perform spatial and motion modeling on the heatmap pooled features, facilitating deeper integration with other modalities via generating spatiotemporal features. First, we employ a transform module $\mathcal{T}$ to convert the raw heatmap pooled features $\textbf{\textit{F}}_{\text{p}}$ into spatial features $\textbf{\textit{F}}_{\text{s}}$ and motion features $\textbf{\textit{F}}_{\text{m}}$. The SMCLM fully leverages the spatial and motion information inherent in the heatmap pooled features by converting the features into spatial and motion modalities, providing a more comprehensive representation for human action. Below we describes the implementation process of the transform module $\mathcal{T}$ based on \cite{10113233}. Given two heatmap pooled features $\textbf{\textit{F}}_{\text{p}}^{i} = \{c_1^i, c_2^i, \dots, c_n^i\}$ and $\textbf{\textit{F}}_{\text{p}}^{j} = \{c_1^j, c_2^j, \dots, c_n^j\}$ with natural topological connections, the spatial features are transformed into $\textbf{\textit{F}}_{\text{s}}^{ij} = \{c_1^i - c_1^j, c_2^i - c_2^j, \dots, c_n^i - c_n^j\}$. Given two heatmap pooled features $\textbf{\textit{F}}_{\text{tp}}^{i} = \{c_{t1}^i, c_{t2}^i, \dots, c_{tn}^i\}$ and $\textbf{\textit{F}}_{\text{(t+1)p}}^{i} = \{c_{(t+1)1}^i, c_{(t+1)2}^i, \dots, c_{(t+1)n}^i\}$ from two consecutive frames $t$ and $t+1$, the motion features are transformed into $\textbf{\textit{F}}_{\text{m}}^{i} = \textbf{\textit{F}}_{\text{(t+1)p}}^{i} - \textbf{\textit{F}}_{\text{tp}}^{i}$. Then, we use three independent topological models to effectively model the heatmap pooled features, spatial features and motion features. These well-modeled features are then integrated through concatenation and passed to the text refinement modulation module (TRMM), which is formulated as:
\begin{equation}
\textbf{\textit{F}}_{\text{c}} = \mathcal{M}_1(\textbf{\textit{F}}_{\text{p}}) \oplus \mathcal{M}_2(\textbf{\textit{F}}_{\text{s}}) \oplus \mathcal{M}_3(\textbf{\textit{F}}_{\text{m}}),  
  \label{eq:smclm}
\end{equation}
where $\textbf{\textit{F}}_{\text{c}}$ represents the features obtained after concatenation and $\oplus$ represents the concatenation operation. The symbols $\mathcal{M}_1$, $\mathcal{M}_2$ and $\mathcal{M}_3$ denote three independent topological models. Meanwhile, to constrain the effective training of the above three independent topology models, we calculate the cross-entropy loss between the output features of these three models and the ground truth action labels.

\begin{algorithm}[htb]  
  \caption{Text Refinement Modulation Module}
  \label{alg:TRMM}  
  \begin{algorithmic}[1]  
    \REQUIRE  
      Action label $\text{text} \in \mathbb{R}^{N}$, Heatmap pooled features $\textit{F}_{\text{c}} \in \mathbb{R}^{D}$
    \ENSURE  
      Information-rich action description features $\textit{F}_{\text{text}}^{\prime} \in \mathbb{R}^{N \times C^{\prime}}$ 
    \vspace{-\baselineskip}
    \FOR{action label index $i = 1$ in $N$}
        \STATE $\textit{F}_{\text{text}_i} = \mathcal{E}_{t}(\text{text}_i) \in \mathbb{R}^{C}$
    \ENDFOR
    \STATE Obtain text features: $\textit{F}_{\text{text}} \in \mathbb{R}^{N \times C}$   
    \STATE Calculate scale factor and shift factor: \\ $\gamma \leftarrow MLP(\textit{F}_{\text{c}})$, $\beta \leftarrow MLP(\textit{F}_{\text{c}})$
    \STATE Modulate text features: $\psi(\textit{F}_{\text{c}}, \textit{F}_{\text{text}}) \leftarrow \gamma \times \textit{F}_{\text{text}} + \beta$
    \STATE Refine and project features: \\
    $\eta(\textit{F}_{\text{c}}) \leftarrow Tanh(MLP(\textit{F}_{\text{c}}).unsqueeze(-1) - $ \\ $MLP(\textit{F}_{\text{c}}).unsqueeze(-2)) \circ \textit{F}_{\text{c}}$
    \STATE Aggregate features: $\textit{F}_{\text{text}}^{\prime} \leftarrow \textit{F}_{\text{text}} + \eta(\textit{F}_{\text{c}}) + \eta(\textit{F}_{\text{c}})$ \\
    \RETURN $\textit{F}_{\text{text}}^{\prime} \in \mathbb{R}^{N \times C^{\prime}}$
  \end{algorithmic}  
\end{algorithm}

\subsection{Text Refinement Modulation Module}
Some previous studies have demonstrated that incorporating text features into human action recognition can significantly enhance performance \cite{xiang2023gap, liu2024mmcl, ActionClip, xu2023language}. However, most previous works directly use the encoded text features of action labels, which suffer from the issue of limited information since action labels are inherently simple words. To address this issue, in our proposed Text Refinement Modulation Module (TRMM), we use the concatenated pooled features $\textbf{\textit{F}}_{\text{c}}$ output from the SMCLM to refine and modulate the text features of action labels. By leveraging the powerful generalization ability of the pretrained text model and the concise heatmap pooled features, we generate action text features with richer semantic information to assist in human action recognition. First, we use a pretrained text encoding model to encode the action texts of all categories (\eg \textit{drink water}, \textit{eat meal}, etc. in the NTU 60 dataset), obtaining text features that describe the action categories by:
\begin{equation}
    \textbf{\textit{F}}_{\text{text}} = \big[\mathcal{E}_{t}(\text{text}_i)\big]_{i=1}^N \in \mathbb{R}^{N \times C},
    \label{eq:enc_text}
\end{equation}
where $\mathcal{E}_{t}$ represents the text encoder. The $N$ represents the number of action categories and the $C$ denotes the channel size of the encoded text features. Then, we modulate and refine the encoded text features $\textbf{\textit{F}}_{\text{text}}$ based on the pooled features $\textbf{\textit{F}}_{\text{c}}$,  which is formulated as:
\begin{equation}
    \textbf{\textit{F}}_{\text{text}}^{\prime} = \textbf{\textit{F}}_{\text{text}} + \psi(\textbf{\textit{F}}_{\text{c}}, \textbf{\textit{F}}_{\text{text}}) + \eta(\textbf{\textit{F}}_{\text{c}}),
    \label{eq:TRMM1}
\end{equation}
\begin{equation}
    \psi(\textbf{\textit{F}}_{\text{c}}, \textbf{\textit{F}}_{\text{text}}) = \gamma\textbf{\textit{F}}_{\text{c}}\textbf{\textit{F}}_{\text{text}} + \beta = \mathcal{P}_{1}(\textbf{\textit{F}}_{\text{c}})\textbf{\textit{F}}_{\text{text}} + \mathcal{P}_2(\textbf{\textit{F}}_{\text{c}}),
    \label{eq:TRMM2}
\end{equation}
\begin{equation}
    \eta(\textbf{\textit{F}}_{\text{c}}) = \big[\mathcal{P}_3(\textbf{\textit{F}}_{\text{c}}) - \mathcal{P}_4(\textbf{\textit{F}}_{\text{c}})\big] \circ \textbf{\textit{F}}_{\text{c}},
    \label{eq:TRMM3}
\end{equation}
where $\gamma$ and $\beta$ represent the scale factor and shift factor respectively. The $\circ$ represents the element-wise multiplication and the $\mathcal{P}$ represents the projection layer, which is implemented as an $MLP$ network layer in our HP-Net. In detail, each projection layer in TRMM is composed of two Linear layers followed by a ReLU activation function. It is worth emphasizing that the four projection layers used in TRMM are implemented as independent $MLP$ networks, without parameter sharing. To provide a clearer understanding of the workflow of the TRMM, we show its pseudocode in Algorithm \ref{alg:TRMM}. In addition, the information-rich action description features $\textbf{\textit{F}}_{\text{text}}^{\prime}$ obtained through the TRMM can be further fused with RGB video features, enabling more accurate multimodal-based human action recognition.

\begin{table*}[t]
\footnotesize
\caption{Accuracy comparison with state-of-the-art methods. In the NTU RGB+D 60 and NTU RGB+D 120 datasets, we report the Top-1 accuracy, while in the Toyota-Smarthome dataset, we report the mean per-class accuracy. The symbols R, P and D denote the RGB, pose and depth modalities, respectively. The symbol $\star$ denotes the rerunning result of authors' original code.}
\begin{tabular*}{\textwidth}{@{\extracolsep\fill}lrcccccccc@{}}
\toprule
     \multirow{2}{*}{\textbf{Method}} & \multirow{2}{*}{\textbf{Source}} & \multirow{2}{*}{\textbf{Visual Modality}} & \multicolumn{2}{c}{\textbf{NTU RGB+D 60 ($\%$)}} & \multicolumn{2}{c}{\textbf{NTU RGB+D 120 ($\%$)}} & \multicolumn{3}{c}{\textbf{Toyota-Smarthome ($\%$)}}\\
\cmidrule{4-5}\cmidrule{6-7}\cmidrule{8-10}
    & &  & {\textbf{X-Sub}}  & \textbf{X-View} & \textbf{X-Sub} & \textbf{X-Set} & \textbf{CS} & \textbf{CV$_{1}$} & \textbf{CV$_{2}$} \\
\midrule
    VPN \cite{das2020vpn} & ECCV'20 & R+P & 95.5 & 98.0 & 87.8 & 86.3 & 60.8 & 43.8 & 53.5 \\
    VPN++ \cite{das2021vpn++} & TPAMI'21 & R+P & 96.6 & 99.1 & 90.7 & 92.5 & 71.0 & - & 58.1 \\
    TSMF \cite{bruce2021multimodal} & AAAI'21 & R+P & 92.5 & 97.4 & 87.0 & 89.1 & - & - & - \\
    DRDIS \cite{9422825} & TCSVT'22 & R+D & 91.1 & 94.3 & 81.3 & 83.4 & - & - & - \\
    MMNet \cite{MMNet} & TPAMI'22 & R+P & 96.0 & 98.8 & 92.9 & 94.4 & 70.1 & 37.4 & 46.6 \\
    VideoMAE$^\star$ \cite{tong2022videomae} & NIPS‘22 & R & 89.3 & 91.2 & 86.5 & 88.2 & - & - & - \\
    PoseConv3D \cite{PoseC3D} & CVPR'22 & R+P & \underline{97.0} & \underline{99.6} & \underline{95.3} & \underline{96.4} & 53.8 & 21.5 & 33.4 \\
    ViewCon \cite{shah2023multi} & WACV'23 & R+P & 93.7 & 98.9 & 85.6 & 87.5 & - & - & - \\
    STAR-Transformer \cite{ahn2023star} & WACV'23 & R+P & 92.0 & 96.5 & 90.3 & 92.7 & - & - & - \\
    3DA \cite{kim2023cross} & ICCV'23 & R+P & 94.3 & 97.9 & 90.5 & 91.4 & - & - & - \\
    MMCL \cite{liu2024mmcl} & MM’24 & R+P & 93.5 & 97.4 & 90.3 & 91.7 & - & - & - \\
    DVANet \cite{siddiqui2024dvanet} & AAAI'24 & R & 93.4 & 98.2 & 90.4 & 91.6 & - & - & - \\
    $\pi$-ViT \cite{reilly2024pivit} & CVPR'24 & R+P & 96.3 & 99.0 & 95.1 & 96.1 & \underline{73.1} & \underline{55.6} & \underline{65.0} \\
    VA-AR \cite{wei2025va} & AAAI'25 & P & 93.1 & 97.2 & 90.3 & 91.5 & - & - & - \\
    Hulk \cite{10930828} & TPAMI'25 & P & 94.3 & - & - & - & - & - & - \\ 
    LA-GCN \cite{xu2023language} & TMM'25 & P & 93.5 & 97.2 & 90.7 & 91.8 & - & - & - \\ 
    Heterogeneous \cite{Wang_2025_CVPR} & CVPR'25 & R+P & 87.8 & 93.7 & 78.9 & 82.2 & - & - & - \\
    ProtoGCN \cite{liu2024revealing} & CVPR'25 & P & 93.8 & 97.8 & 90.9 & 92.2 & - & - & - \\
\midrule
    \textbf{HP-Net (1s)} & \textbf{Ours} & R & \textbf{97.0} & \textbf{99.0} & \textbf{96.3} & \textbf{96.5} & \textbf{81.3} & \textbf{51.7} & \textbf{66.4} \\ 
    \textbf{HP-Net} & \textbf{Ours} & R & \textbf{97.9} & \textbf{99.7} & \textbf{96.8} & \textbf{97.5} & \textbf{81.6} & \textbf{57.9} & \textbf{67.4} \\ 
\bottomrule
\label{tab:sota}
\end{tabular*}
\end{table*}

\subsection{Multimodal Fusion and Loss Function}
In video-based human action recognition, previous methods have demonstrated that integrating features from multiple modalities leads to more accurate recognition performance \cite{MMNet, PoseC3D, reilly2024pivit}. Therefore, our proposed HP-Net also fuses RGB video features, textual features and heatmap pooled features to achieve multimodal action recognition. Specifically, in addition to utilizing the compact heatmap pooled features produced by the FPM and the semantically rich text features generated by the TRMM, we also extract features from RGB video frames using a pre-trained image encoder. In detail, given an RGB video stream $\textbf{\textit{U}} = \{u_1, u_2, \dots, u_n\}$ with n frames, we extract the RGB video features by:
\begin{equation}
    \textbf{\textit{F}}_{\text{u}} = \mathcal{E}_{u}\big[\mathcal{U}_n^m(\mathcal{D}(u_i) \stackrel{(\text{h}^\prime, \text{w}^\prime)}{\mapsto} u_i^{\prime}) \big], \quad \forall i \in \{1, 2, \dots, n\},
    \label{eq:detect}
\end{equation}
where $\mathcal{D}$ denotes a human detector. Inspired by the top-down pose estimation, we use a human detector $\mathcal{D}$ to filter out most ambient noise and focus on the human body, which has already been validated in some previous works\cite{liu2024mmcl}. The $\mathcal{D}(u_i) \stackrel{(\text{h}^\prime, \text{w}^\prime)}{\mapsto} u_i^{\prime}$ represents the resizing operation applied to the detected human frames to save computational resources and adapt to the downstream RGB visual encoder. Then, we sample \textit{m} frames from \textit{n} detected human frames using $\mathcal{U}_n^m$ and input them into the visual encoder $\mathcal{E}_{u}$ to obtain the RGB video features. Finally, we fuse the text features $\textbf{\textit{F}}_{\text{text}}^{\prime}$ refined and modulated by the heatmap pooled features $\textbf{\textit{F}}_{\text{c}}$ with the video features $\textbf{\textit{F}}_{\text{u}}$ to obtain the classification scores $\textbf{\textit{S}}$ for each action category. In summary, the fusion of multi-modal features in our proposed HP-Net can be outlined as the following mapping process:
\begin{gather}
    \textbf{\textit{U}} \stackrel{\text{FPM}}{\mapsto} \textbf{\textit{F}}_{\text{p}} \stackrel{\text{SMCLM}}{\mapsto} \textbf{\textit{F}}_{\text{c}} \stackrel{\text{TRMM}}{\mapsto} \textbf{\textit{F}}_{\text{text}}^{\prime}, \\
    \textbf{\textit{S}} = \textbf{\textit{F}}_{\text{text}}^{\prime} \circ \textbf{\textit{F}}_{\text{u}}.
\end{gather}
In our proposed Heatmap Pooling Network (HP-Net), we calculate the cross-entropy loss between the ground truth action labels and the classification scores as the classification loss by:
\begin{equation}
  \mathcal{L}_\text{cls} = \mathcal{L}_\text{CE}(\textbf{\textit{Y}}, \textbf{\textit{S}}) = -\sum_{i}^{N} \textbf{\textit{Y}}^{i}log(\textbf{\textit{S}}^i),
  \label{eq:cls}
\end{equation}
where $N$ is the number of samples in a batch and $\textbf{\textit{Y}}^{i}$ is the one-hot presentation of the ground truth action label about data sample $\textit i$. The $\textbf{\textit{S}}^i$ is the classification score of data sample $\textit i$. Besides, to constrain the effective training of three independent topology models $\mathcal{M}_1$, $\mathcal{M}_2$ and $\mathcal{M}_3$ in the Spatial-motion Co-learning Module (SMCLM), we also calculate the cross-entropy loss between the output features of these three models and the ground truth action labels as the auxiliary loss. Therefore, the training loss function for the entire Heatmap Pooling Network (HP-Net) is defined as:
\begin{equation}
  \mathcal{L}_\text{loss} = \mathcal{L}_\text{cls} + \lambda_1\mathcal{L}_\text{p} + \lambda_2\mathcal{L}_\text{s} + \lambda_3\mathcal{L}_\text{m},
  \label{eq:loss}
\end{equation}
where $\mathcal{L}_\text{p} = \mathcal{L}_\text{CE}(\textbf{\textit{Y}}, \varphi(\textbf{\textit{F}}_{\text{p}}^{\prime}))$,  $\mathcal{L}_\text{s} = \mathcal{L}_\text{CE}(\textbf{\textit{Y}}, \varphi(\textbf{\textit{F}}_{\text{s}}^{\prime}))$ and $\mathcal{L}_\text{m} = \mathcal{L}_\text{CE}(\textbf{\textit{Y}}, \varphi(\textbf{\textit{F}}_{\text{m}}^{\prime}))$. The $\varphi$ represents the classification head used to convert features into classification scores. The $\lambda_1$, $\lambda_2$ and $\lambda_3$ represent three hyper-parameters that regulate the weights of the different auxiliary loss terms.

\begin{table}[t]
\footnotesize
\caption{Accuracy comparison with state-of-the-art methods on the UAV-Human dataset. The symbols R and P denote the RGB and pose modalities, respectively. The symbol $\star$ denotes the rerunning result of authors' original code.}
\begin{tabular}{lrccc}
\toprule
\multirow{2}{*}{\textbf{Method}} & \multirow{2}{*}{\textbf{Source}} & \textbf{Visual} & \multicolumn{2}{c}{\textbf{UAV-Human ($\%$)}}\\
& & \textbf{Modality} & \textbf{CSv1} & \textbf{CSv2} \\
\midrule
ST-GCN \cite{yan2018spatial} & AAAI'18 & P & 30.3 & 56.1 \\
DGNN \cite{shi2019skeleton} & CVPR'19 & P & 29.9 & - \\
2s-AGCN \cite{shi2019two} & CVPR'19 & P & 34.8 & 66.7 \\
HARD-Net \cite{li2020hard} & ECCV'20 & P & 37.0 & - \\
Shift-GCN \cite{shiftgcn2020} & CVPR'20 & P & 38.0 & 67.0 \\
CTR-GCN \cite{Chen_2021_ICCV} & ICCV'21 & P & 45.6 & 73.5 \\
FAR \cite{kothandaraman2022far} & ECCV'21 & R & 39.1 & - \\
FR-AGCN \cite{hu2022forward} & Neuroc'22 & P & 44.0 & 69.5 \\
ACFL-CTR \cite{wang2022skeleton} & ACM MM'22 & P & 45.3 & - \\
MixFormer \cite{xin2023skeleton} & ACM MM'23 & P & 47.2 & 73.5 \\
AZTR \cite{wang2023aztr} & ICRA'23 & R & 47.4 & - \\
TD-GCN \cite{10113233} & TMM'24 & P & 45.4 & 72.9 \\
HDBN \cite{liu2024HDBN} & ICME'24 & P & 48.0 & \underline{75.4} \\
MITFAS \cite{xian2024mitfas} & WACV'24 & R & \underline{50.8} & - \\
STRN \cite{wang2025structural} & TIM'25 & P & 45.3 & 72.0 \\ 
2D$^3$-SkelAct \cite{zhang2025robust} & TCSVT'25 & P & 32.5 & 56.2 \\
ProtoGCN$^\star$ \cite{liu2024revealing} & CVPR'25 & P & 46.1 & 73.5 \\
\midrule
\textbf{HP-Net} (1s) & \textbf{Ours} & R & \textbf{52.0} & \textbf{77.8} \\
\textbf{HP-Net} & \textbf{Ours} & R & \textbf{53.6} & \textbf{80.2} \\
\bottomrule
\label{tab:uav}
\end{tabular}
\end{table}

\begin{table}[t]
\footnotesize
\caption{Comparison between 2D pose and heatmap pooled features. We compared the performance under four modalities: Joint (J), Bone (B), Joint motion (JM) and Bone motion (BM). Ensemble refers to combining the classification scores of the above four modalities.}
\centering
\begin{tabular}{cccc}
\hline
\textbf{Methods} & \textbf{Visual Modality} & \textbf{Acc.} ($\%$) \\\hline
CTR-GCN \cite{Chen_2021_ICCV} & J 2D Pose & 83.0 \\
\textbf{CTR-GCN} & \textbf{J Pooled Feature} & $\textbf{86.1}^{\uparrow\textbf{3.1}}$  \\
\midrule
CTR-GCN & JM 2D Pose & 80.3 \\
\textbf{CTR-GCN} & \textbf{JM Pooled Feature} & $\textbf{83.2}^{\uparrow\textbf{2.9}}$  \\
\midrule
CTR-GCN & B 2D Pose & 83.9 \\
\textbf{CTR-GCN} & \textbf{B Pooled Feature} & $\textbf{85.3}^{\uparrow\textbf{1.4}}$  \\
\midrule
CTR-GCN & BM 2D Pose & 80.0 \\
\textbf{CTR-GCN} & \textbf{BM Pooled Feature} & $\textbf{82.0}^{\uparrow\textbf{2.0}}$  \\
\midrule
CTR-GCN & Ensemble & 86.2 \\
\textbf{CTR-GCN} & \textbf{Ensemble} & $\textbf{88.7}^{\uparrow\textbf{2.5}}$  \\
\midrule
Skeleton-MixFormer \cite{xin2023skeleton} & J 2D Pose & 82.9 \\
\textbf{Skeleton-MixFormer} & \textbf{J Pooled Feature} & $\textbf{85.9}^{\uparrow\textbf{3.0}}$  \\
\midrule
Skeleton-MixFormer & JM 2D Pose & 79.2 \\
\textbf{Skeleton-MixFormer} & \textbf{JM Pooled Feature} & $\textbf{82.0}^{\uparrow\textbf{2.8}}$  \\
\midrule
Skeleton-MixFormer & B 2D Pose & 84.0 \\
\textbf{Skeleton-MixFormer} & \textbf{B Pooled Feature} & $\textbf{84.3}^{\uparrow\textbf{0.3}}$  \\
\midrule
Skeleton-MixFormer & BM 2D Pose & 80.1 \\
\textbf{Skeleton-MixFormer} & \textbf{BM Pooled Feature} & $\textbf{81.0}^{\uparrow\textbf{0.9}}$  \\
\midrule
Skeleton-MixFormer & Ensemble & 86.2 \\
\textbf{Skeleton-MixFormer} & \textbf{Ensemble} & $\textbf{88.4}^{\uparrow\textbf{2.2}}$  \\
\hline
\end{tabular}
\label{tab:FPM}
\end{table}

\begin{table}[t]
\footnotesize
\centering
    \caption{Comparison of different modalities using the same model backbone. Here we use the same pose estimation model and action recognition backbone to ensure a fair comparison.}
    \begin{tabular}{ccc}
    \hline
    \textbf{Visual Modality} & \textbf{Source} & \textbf{Acc.} ($\%$) \\\hline
    2D Pose Data & Pose Estimation \cite{li2022simcc} & 83.0 \\
    3D Pose Data & Kinect Sensor \cite{8713892} & 85.0 \\
    Heatmap Coordinate Vector & Pose Estimation \cite{li2022simcc} & 83.9 \\
    \textbf{Heatmap Pooled Feature} & \textbf{Ours} & \textbf{86.1} \\\hline
    \end{tabular}
    \centering
    \label{tab:modality}
\end{table}

\begin{table*}[t]
\footnotesize
\caption{Accuracy Improvement of Heatmap Pooled Features Over 2D/3D Pose Data for Some Action Categories. The blue and red fonts indicate the improvement rates relative to 2D pose and 3D pose, respectively.}
\centering
\begin{tabular}{lccccccc} 
\toprule
\textbf{Action Label} & \textbf{2}  & \textbf{3} & \textbf{4} & \textbf{5} & \textbf{11} & \textbf{12} & \textbf{18}\\
\midrule
\textbf{Acc.} (2D Pose data) ($\%$) & 80.7 & 83.5 & 87.2 & 85.8 & 56.4 & 64.0 & 93.8 \\
\textbf{Acc.} (3D Pose data) ($\%$) & 72.7 & 83.5 & 86.4 & 84.4 & 64.5 & 53.7 & 91.6 \\
\textbf{Acc. (Pooled Feature)} ($\%$) & 82.9$^{\uparrow\textcolor{blue}{\textbf{2.2}}}$$^{\uparrow\textcolor{red}{\textbf{10.2}}}$ & 90.1$^{\uparrow\textcolor{blue}{\textbf{6.6}}}$$^{\uparrow\textcolor{red}{\textbf{6.6}}}$ & 89.4$^{\uparrow\textcolor{blue}{\textbf{2.2}}}$$^{\uparrow\textcolor{red}{\textbf{3.0}}}$ & 90.9$^{\uparrow\textcolor{blue}{\textbf{5.1}}}$$^{\uparrow\textcolor{red}{\textbf{6.5}}}$ & 70.3$^{\uparrow\textcolor{blue}{\textbf{13.9}}}$$^{\uparrow\textcolor{red}{\textbf{5.8}}}$ & 71.3$^{\uparrow\textcolor{blue}{\textbf{7.3}}}$$^{\uparrow\textcolor{red}{\textbf{17.6}}}$ & 96.0$^{\uparrow\textcolor{blue}{\textbf{2.2}}}$$^{\uparrow\textcolor{red}{\textbf{4.4}}}$ \\
\midrule
\textbf{Action Label} & \textbf{29}  & \textbf{30} & \textbf{31} & \textbf{37} & \textbf{44} & \textbf{47} & \textbf{50}\\
\midrule
\textbf{Acc.} (2D Pose data) ($\%$) & 53.1 & 78.9 & 71.4 & 86.2 & 87.7 & 90.2 & 86.5 \\
\textbf{Acc.} (3D Pose data) ($\%$) & 61.1 & 70.5 & 76.1 & 87.3 & 78.3 & 86.2 & 87.6 \\
\textbf{Acc. (Pooled Feature)} ($\%$) & 73.5$^{\uparrow\textcolor{blue}{\textbf{20.4}}}$$^{\uparrow\textcolor{red}{\textbf{12.4}}}$ & 84.0$^{\uparrow\textcolor{blue}{\textbf{5.1}}}$$^{\uparrow\textcolor{red}{\textbf{13.5}}}$ & 80.8$^{\uparrow\textcolor{blue}{\textbf{9.4}}}$$^{\uparrow\textcolor{red}{\textbf{4.7}}}$ & 94.6$^{\uparrow\textcolor{blue}{\textbf{8.4}}}$$^{\uparrow\textcolor{red}{\textbf{7.3}}}$ & 92.0$^{\uparrow\textcolor{blue}{\textbf{4.3}}}$$^{\uparrow\textcolor{red}{\textbf{13.7}}}$ & 95.7$^{\uparrow\textcolor{blue}{\textbf{5.5}}}$$^{\uparrow\textcolor{red}{\textbf{9.5}}}$ & 91.6$^{\uparrow\textcolor{blue}{\textbf{5.1}}}$$^{\uparrow\textcolor{red}{\textbf{4.0}}}$  \\
\midrule
\textbf{Action Label} & \textbf{54} & \textbf{56} & \textbf{57} & \textbf{58} & \textbf{61} & \textbf{62} & \textbf{67}\\
\midrule
\textbf{Acc.} (2D Pose data) ($\%$) & 90.6 & 89.9 & 88.0 & 91.7 & 89.8 & 90.3 & 71.0 \\
\textbf{Acc.} (3D Pose data) ($\%$) & 90.2 & 87.7 & 92.0 & 95.7 & 92.7 & 89.8 & 78.9 \\
\textbf{Acc. (Pooled Feature)} ($\%$) & 93.5$^{\uparrow\textcolor{blue}{\textbf{2.9}}}$$^{\uparrow\textcolor{red}{\textbf{3.3}}}$ & 92.4$^{\uparrow\textcolor{blue}{\textbf{2.5}}}$$^{\uparrow\textcolor{red}{\textbf{4.7}}}$ & 95.3$^{\uparrow\textcolor{blue}{\textbf{7.3}}}$$^{\uparrow\textcolor{red}{\textbf{3.3}}}$ & 98.9$^{\uparrow\textcolor{blue}{\textbf{7.2}}}$$^{\uparrow\textcolor{red}{\textbf{3.2}}}$ & 94.8$^{\uparrow\textcolor{blue}{\textbf{5.0}}}$$^{\uparrow\textcolor{red}{\textbf{2.1}}}$ & 92.9$^{\uparrow\textcolor{blue}{\textbf{2.6}}}$$^{\uparrow\textcolor{red}{\textbf{3.1}}}$ & 85.9$^{\uparrow\textcolor{blue}{\textbf{14.9}}}$$^{\uparrow\textcolor{red}{\textbf{7.0}}}$ \\
\midrule
\textbf{Action Label} & \textbf{68}  & \textbf{71} & \textbf{75} & \textbf{79} & \textbf{86} & \textbf{93} & \textbf{105}\\
\midrule
\textbf{Acc.} (2D Pose data) ($\%$) & 81.9 & 44.2 & 67.8 & 80.3 & 76.0 & 77.8 & 68.7 \\
\textbf{Acc.} (3D Pose data) ($\%$) & 79.8 & 45.7 & 68.9 & 76.5 & 78.6 & 80.0 & 64.5 \\
\textbf{Acc. (Pooled Feature)} ($\%$) & 85.7$^{\uparrow\textcolor{blue}{\textbf{3.8}}}$$^{\uparrow\textcolor{red}{\textbf{5.9}}}$ & 53.9$^{\uparrow\textcolor{blue}{\textbf{9.7}}}$$^{\uparrow\textcolor{red}{\textbf{8.2}}}$ & 78.9$^{\uparrow\textcolor{blue}{\textbf{11.1}}}$$^{\uparrow\textcolor{red}{\textbf{10.0}}}$ & 86.3$^{\uparrow\textcolor{blue}{\textbf{5.9}}}$$^{\uparrow\textcolor{red}{\textbf{9.7}}}$ & 81.9$^{\uparrow\textcolor{blue}{\textbf{5.9}}}$$^{\uparrow\textcolor{red}{\textbf{3.3}}}$ & 82.3$^{\uparrow\textcolor{blue}{\textbf{4.5}}}$$^{\uparrow\textcolor{red}{\textbf{2.3}}}$ & 83.3$^{\uparrow\textcolor{blue}{\textbf{14.6}}}$$^{\uparrow\textcolor{red}{\textbf{18.8}}}$ \\
\midrule
\textbf{Action Label} & \textbf{106}  & \textbf{108} & \textbf{110} & \textbf{111} & \textbf{116} & \textbf{117} & \textbf{118}\\
\midrule
\textbf{Acc.} (2D Pose data) ($\%$) & 60.7 & 88.5 & 72.5 & 92.5 & 95.0 & 90.1 & 90.3 \\
\textbf{Acc.} (3D Pose data) ($\%$) & 62.4 & 91.8 & 70.8 & 93.0 & 91.7 & 91.8 & 94.4 \\
\textbf{Acc. (Heatmap Pooling)} ($\%$) & 66.1$^{\uparrow\textcolor{blue}{\textbf{5.4}}}$$^{\uparrow\textcolor{red}{\textbf{3.7}}}$ & 94.8$^{\uparrow\textcolor{blue}{\textbf{6.3}}}$$^{\uparrow\textcolor{red}{\textbf{3.0}}}$ & 74.8$^{\uparrow\textcolor{blue}{\textbf{2.3}}}$$^{\uparrow\textcolor{red}{\textbf{4.0}}}$ & 96.9$^{\uparrow\textcolor{blue}{\textbf{4.4}}}$$^{\uparrow\textcolor{red}{\textbf{3.9}}}$ & 98.6$^{\uparrow\textcolor{blue}{\textbf{3.6}}}$$^{\uparrow\textcolor{red}{\textbf{6.9}}}$ & 95.7$^{\uparrow\textcolor{blue}{\textbf{5.6}}}$$^{\uparrow\textcolor{red}{\textbf{3.8}}}$ & 97.6$^{\uparrow\textcolor{blue}{\textbf{7.3}}}$$^{\uparrow\textcolor{red}{\textbf{3.2}}}$ \\
\bottomrule
\end{tabular}
\label{tab:action_acc}
\end{table*}

\begin{figure*}[t]
  \centering
   \includegraphics[width=1.0\linewidth]{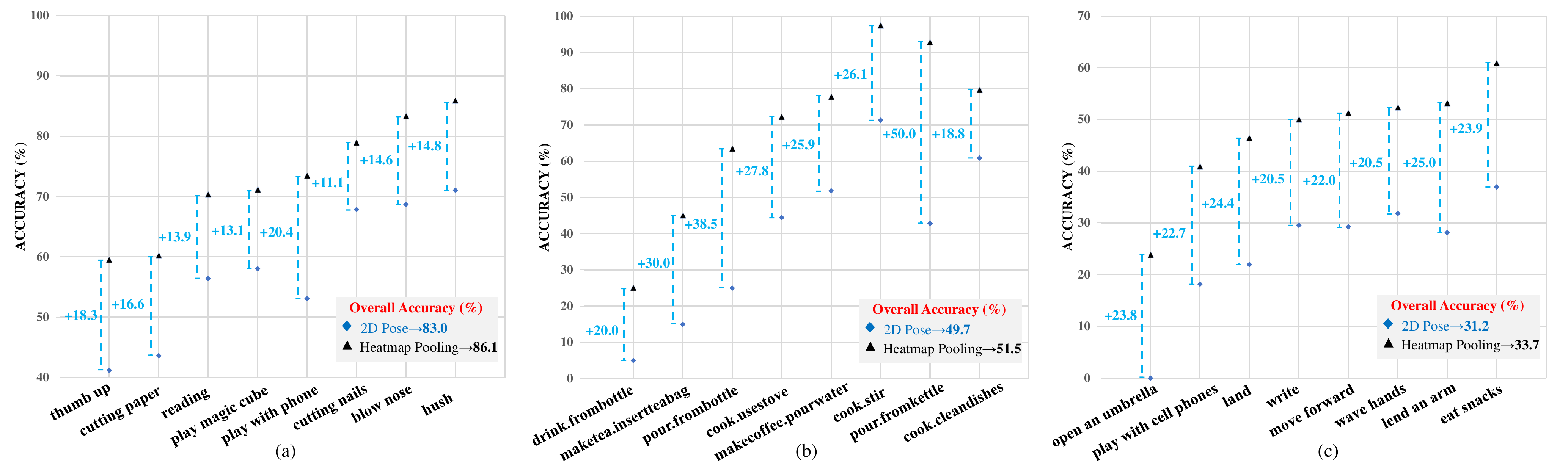}
   \caption{Visualization of the recognition accuracy for some action categories. (a) The top-8 action categories where the use of heatmap pooled features outperforms 2D pose in the NTU 120 dataset. (b) The top-8 action categories where the use of heatmap pooled features outperforms 2D pose in the Toyota-Smarthome dataset. (c) The top-8 action categories where the use of heatmap pooled features outperforms 2D pose in the UAV-Human dataset.}
   \label{fig:acc}
\end{figure*}

\begin{figure*}[t]
  \centering
   \includegraphics[width=1.0\linewidth]{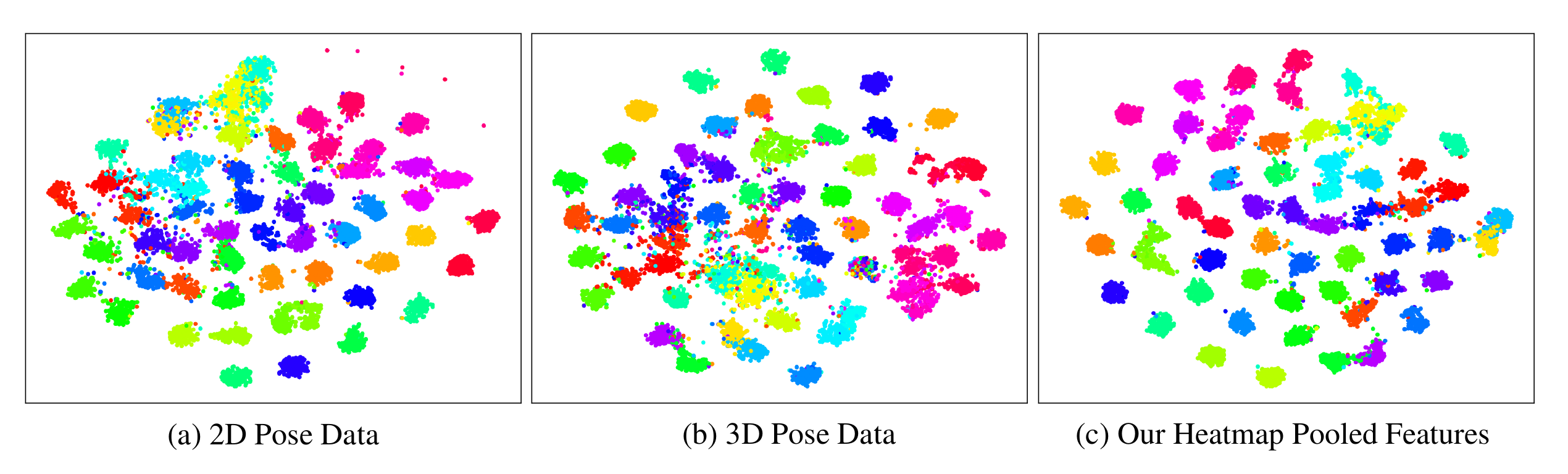}
   \caption{T‐distributed stochastic neighbor embedding (T‐SNE) visualization of features. We use the same network backbone to model three types of data (\ie 2D pose data, 3D pose data and heatmap pooled features) and visualize the modeled features. The 2D pose data and heatmap pooled features come from the same pose estimation network (\ie SimCC \cite{li2022simcc}), while the 3D pose data is provided by the original NTU-60 dataset.}
   \label{fig:tsne}
\end{figure*}

\begin{table}[t]
\footnotesize
\centering
    \caption{Comparison of accuracy when obtain heatmap pooled features using different human pose estimation methods. The HP stands for heatmap pooled features.}
    \begin{tabular}{cccc}
    \hline
    \textbf{Modality} & \textbf{Source} & \textbf{Backbone} & \textbf{Acc.} ($\%$) \\\hline
    HP & HR-Net \cite{wang2020deep} & CTR-GCN \cite{Chen_2021_ICCV}  & 85.3 \\
    HP & SimCC \cite{li2022simcc} & CTR-GCN & 86.1 \\\hline
    HP & HR-Net & Skeleton-MixFormer \cite{xin2023skeleton} & 84.0 \\
    HP & SimCC & Skeleton-MixFormer & 85.9 \\\hline
    HP+RGB+Text & HR-Net & HP-Net (Ours) & 95.8 \\
    HP+RGB+Text & SimCC & HP-Net (Ours) & 96.3 \\\hline
    \end{tabular}
    \centering
    \label{tab:HPE_Acc}
\end{table}

\begin{table}[t]
\footnotesize
\centering
    \caption{Comparison of the FPM when pooling on heatmaps at different scales. $h$, $w$ and $c$ represent the height, width and channel size of the heatmaps, while $n$ denotes the number of joints used in the  feedback pooling module (FPM).}
    \begin{tabular}{cccc}
    \hline
    \textbf{Heatmap} & \textbf{Heatmap Shape} & \textbf{Feature Shape}  & \textbf{Acc.} ($\%$) \\\hline
    $H_{1}$ & [$c_{1}$, $h_{1}$, $w_{1}$] & [$n$, $c_{1}$] & 81.4 \\
    $H_{2}$ & [$c_{2}$, $h_{2}$, $w_{2}$] & [$n$, $c_{2}$] & \textbf{86.2} \\
    $H_{3}$ & [$c_{3}$, $h_{3}$, $w_{3}$] & [$n$, $c_{3}$] & 85.8 \\\hline
    \end{tabular}
    \centering
    \label{tab:scale}
\end{table}

\begin{figure}[t]
  \centering
   \includegraphics[width=0.9\linewidth]{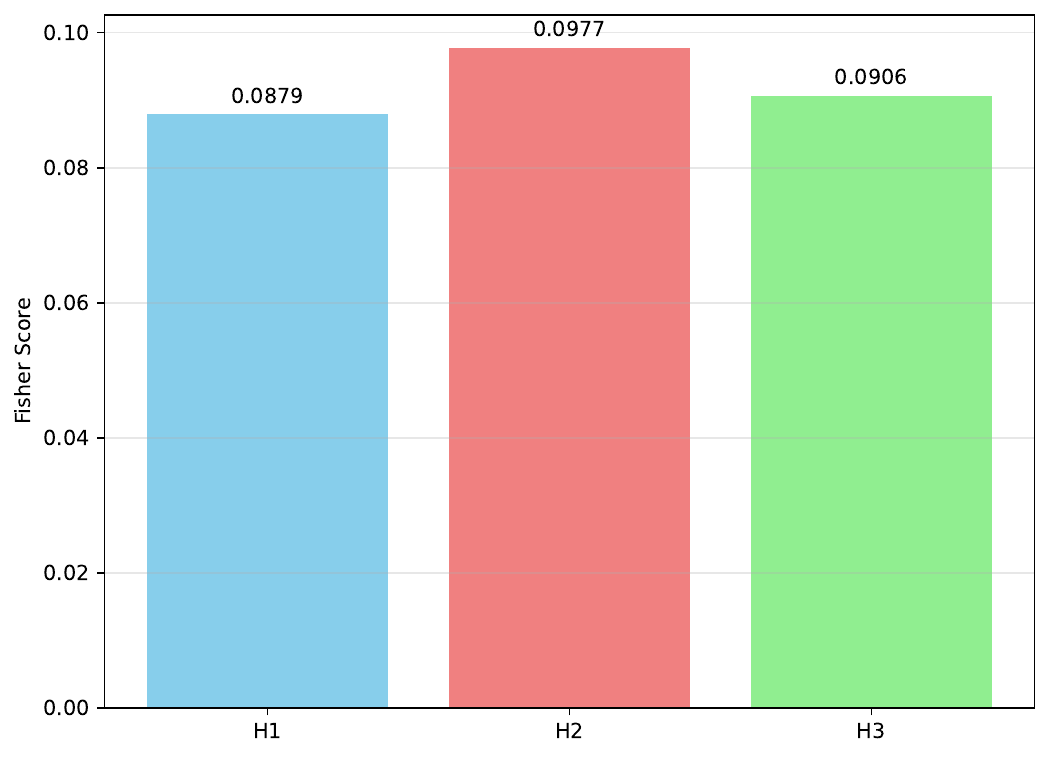}
   \caption{Visualization of the Fisher scores of heatmaps at different scales. A higher Fisher score indicates greater feature discriminability.}
   \label{fig:fisher}
\end{figure}

\begin{table}[t]
\footnotesize
\centering
    \caption{Comparison of the FPM when pooling on heatmaps at different stages.}
    \begin{tabular}{ccccc}
    \hline
    \textbf{Heatmap} & \textbf{Stages} & \textbf{Heatmap Shape} & \textbf{Backbone}  & \textbf{Acc.} ($\%$) \\\hline
    $H_{2}^1$ & Stage1 & [$c_{2}$, $h_{2}$, $w_{2}$] & CTR-GCN & 69.4 \\
    $H_{2}^2$ & Stage2 & [$c_{2}$, $h_{2}$, $w_{2}$] & CTR-GCN & 75.8 \\
    $H_{2}^3$ & Stage3 & [$c_{2}$, $h_{2}$, $w_{2}$] & CTR-GCN & \textbf{86.2} \\
    \hline
    $H_{2}^1$ & Stage1 & [$c_{2}$, $h_{2}$, $w_{2}$] & HP-Net & 95.8 \\
    $H_{2}^2$ & Stage2 & [$c_{2}$, $h_{2}$, $w_{2}$] & HP-Net & 95.9 \\
    $H_{2}^3$ & Stage3 & [$c_{2}$, $h_{2}$, $w_{2}$] & HP-Net & \textbf{96.3} \\
    \hline
    \end{tabular}
    \centering
    \label{tab:stage}
\end{table}

\begin{table}[t]
\footnotesize
\centering
    \caption{Comparison of the FPM when performing pooling over heatmaps using regions of varying sizes.}
    \begin{tabular}{ccccc}
    \hline
    \textbf{Heatmap} & \textbf{Stages} & \textbf{Region Size} & \textbf{Backbone}  & \textbf{Acc.} ($\%$) \\\hline
    $H_{3}$  & Stage3 & $1\times1$ & CTR-GCN & \textbf{86.2} \\
    $H_{3}$  & Stage3 & $3\times3$ & CTR-GCN & 82.0 \\
    $H_{3}$  & Stage3 & $5\times5$ & CTR-GCN & 81.9 \\
    \hline
    $H_{3}$  & Stage3 & $1\times1$ & HP-Net & \textbf{96.3} \\
    $H_{3}$  & Stage3 & $3\times3$ & HP-Net & 96.2 \\
    $H_{3}$  & Stage3 & $5\times5$ & HP-Net & 96.1 \\
    \hline
    \end{tabular}
    \centering
    \label{tab:region}
\end{table}

\begin{table}[t]
\footnotesize
\centering
    \caption{Comparison in parameters and computation cost when modeling heatmap features.}
    \begin{tabular}{cccc}
    \hline
    \textbf{Method} & \textbf{Backbone} & \textbf{Param.}  & \textbf{FLOPs} \\\hline
    Heatmap Evolution \cite{Liu_2018_CVPR} & 2D CNN & 140.1M & 19.6G \\
    Heatmap Volume \cite{PoseC3D} & 3D CNN & 2.0M & 20.6G \\
    \textbf{Heatmap Pooling (Ours)} & \textbf{GCN} & \textbf{1.4M} & \textbf{1.8G} \\\hline
    \end{tabular}
    \centering
    \label{tab:param}
\end{table}

\begin{table}[t]
\footnotesize
\centering
    \caption{Comparison of pose-guided regional sampling across different modalities. The HP stands for heatmap pooled features.}
    \begin{tabular}{cccc}
    \hline
    \textbf{Method} & \textbf{Modality} & \textbf{Backbone}  & \textbf{Acc.} ($\%$) \\\hline
   RGB Regional Sampling & RGB+Pose & CTR-GCN & 84.8 \\
    \textbf{Heatmap Regional Sampling} & \textbf{HP} & \textbf{CTR-GCN} & \textbf{86.1} \\\hline
    \end{tabular}
    \centering
    \label{tab:pose-guided}
\end{table}

\begin{table}[t]
\footnotesize
\centering
    \caption{Comparison of accuracy when transfer heatmap pooled features using different modules. The HP stands for heatmap pooled features.}
    \begin{tabular}{ccc}
    \hline
    \textbf{Method} & \textbf{Modality} & \textbf{Acc.} ($\%$) \\\hline
    Baseline & Text+RGB  & 95.3 \\
    w/ FPM+Cross Attention & Text+RGB+HP  & $\textbf{95.4}^{\uparrow\textbf{0.1}}$ \\
    w/ FPM+FiLM & Text+RGB+HP  & $\textbf{95.6}^{\uparrow\textbf{0.3}}$ \\
    w/ FPM+TRMM & Text+RGB+HP & $\textbf{95.7}^{\uparrow\textbf{0.4}}$ \\
    w/ FPM+TRMM+SMCLM & Text+RGB+HP & $\textbf{96.3}^{\uparrow\textbf{1.0}}$ \\\hline
    \end{tabular}
    \centering
    \label{tab:TRMM+SMCLM}
\end{table}

\begin{table}[t]
\footnotesize
\centering
    \caption{Comparison of heatmap pooling network with and without using a human detector.}
    \begin{tabular}{ccc}
    \hline
    \textbf{Benchmark} & \textbf{Setting} & \textbf{Acc.} ($\%$) \\\hline
    NTU 120 X-Sub & W/o Human Detector & 87.0 \\
    NTU 120 X-Sub & W/ Human Detector & $\textbf{96.3}^{\uparrow\textbf{9.3}}$ \\\hline
    NTU 120 X-Set & W/o Human Detector & 88.0 \\
    NTU 120 X-Set & W/ Human Detector & $\textbf{96.5}^{\uparrow\textbf{8.5}}$ \\\hline
    Smarthome CS & W/o Human Detector  & 80.7 \\
    Smarthome CS & W/ Human Detector & $\textbf{81.3}^{\uparrow\textbf{0.6}}$ \\\hline
    \end{tabular}
    \centering
    \label{tab:Detection}
\end{table}

\begin{table}[t]
\footnotesize
\centering
    \caption{Comparison of heatmap pooling network with different hyperparameter settings.}
    \begin{tabular}{cccccc}
    \hline
    \textbf{Method} & \textbf{Benchmark} & \textbf{$\lambda_1$} & \textbf{$\lambda_2$} & \textbf{$\lambda_3$} & \textbf{Acc.} ($\%$) \\\hline
    HP-Net & NTU 120 X-Sub & 0.1 & 0.1 & 0.1 & 96.1 \\
    HP-Net & NTU 120 X-Sub & 0.5 & 0.5 & 0.5 & 96.2 \\
    HP-Net & NTU 120 X-Sub & 1.0 & 1.0 & 1.0 & \textbf{96.3} \\
    HP-Net & NTU 120 X-Sub & 2.0 & 2.0 & 2.0 & 96.1 \\\hline
    \end{tabular}
    \centering
    \label{tab:lambda}
\end{table}

\begin{table}[t]
\footnotesize
\centering
    \caption{Comparison of recognition accuracy for action samples under extreme conditions. The HP stands for heatmap pooled features.}
    \begin{tabular}{ccc}
    \hline
    \textbf{Method} & \textbf{Modality} & \textbf{Acc.} ($\%$) \\\hline
    Baseline & Text+RGB & 42.3 \\
    w/ FPM+TRMM+SMCLM & Text+RGB+HP & $\textbf{51.5}^{\uparrow\textbf{9.2}}$ \\\hline
    \end{tabular}
    \centering
    \label{tab:uav_extreme}
\end{table}

\begin{table*}[t]
\footnotesize
\caption{Performance improvements by aggregating the results of multi-stream ensemble compared with the top-10 accurate and confused action samples of the HP-Net (1s). The HP-Net (1s) represents $l_5$ in Tables \ref{tab:Ensemble_NTU} and \ref{tab:Ensemble_SH}, while HP-Net (ms) represents $l1$+$l2$+$l3$+$l4$+$l5$ in Tables \ref{tab:Ensemble_NTU} and \ref{tab:Ensemble_SH}.}
\centering
\begin{tabular}{clcclcc} %
\toprule
\textbf{Benchmark} & \textbf{Accurate Actions}  & \textbf{HP-Net (1s)} ($\%$) & \textbf{HP-Net (ms)} ($\%$) & \textbf{Confused Actions} & \textbf{HP-Net (1s)} ($\%$) & \textbf{HP-Net (ms)} ($\%$) \\
\hline
 & carry object & 100.0 & 100.0 & nausea/vomiting & 84.4 & 86.9$^{\uparrow\textcolor{blue}{\textbf{2.5}}}$ \\ 
 & high-five & 100.0 & 100.0 & sneeze/cough & 89.5 & 89.5 \\  
 & cheers and drink & 100.0 & 100.0 & rub two hands & 89.9 & 91.3$^{\uparrow\textcolor{blue}{\textbf{1.4}}}$ \\ 
 & cross arms & 100.0 & 100.0 & writing & 91.2 & 93.4$^{\uparrow\textcolor{blue}{\textbf{2.2}}}$ \\ 
NTU 120 & run on the spot & 100.0 & 100.0 & put on a shoe & 91.2 & 93.0$^{\uparrow\textcolor{blue}{\textbf{1.8}}}$ \\
X-Sub & squat down & 100.0 & 100.0 & clapping & 91.6 & 93.0$^{\uparrow\textcolor{blue}{\textbf{1.4}}}$ \\ 
 & follow & 99.8 & 100.0$^{\uparrow\textcolor{blue}{\textbf{0.2}}}$ & eat meal & 91.6 & 92.0$^{\uparrow\textcolor{blue}{\textbf{0.4}}}$ \\
 & rock-paper-scissors & 99.7 & 99.8$^{\uparrow\textcolor{blue}{\textbf{0.1}}}$ & take off a shoe & 93.1 & 94.9$^{\uparrow\textcolor{blue}{\textbf{1.8}}}$ \\ 
 & put on bag & 99.7 & 99.8$^{\uparrow\textcolor{blue}{\textbf{0.1}}}$ & hand waving & 93.8 & 94.2$^{\uparrow\textcolor{blue}{\textbf{0.4}}}$ \\ 
 & take off bag & 99.5 & 99.8$^{\uparrow\textcolor{blue}{\textbf{0.3}}}$ & reach into pocket & 95.3 & 96.4$^{\uparrow\textcolor{blue}{\textbf{1.1}}}$ \\ 
\hline 
 & put on a coat & 79.5 & 81.8$^{\uparrow\textcolor{blue}{\textbf{2.3}}}$ & hit someone with something & 21.2 & 27.3$^{\uparrow\textcolor{blue}{\textbf{6.1}}}$ \\ 
 & eat snacks & 76.1 & 76.1 & throw away something & 23.4 & 27.6$^{\uparrow\textcolor{blue}{\textbf{4.2}}}$ \\ 
 & take off a coat & 75.0 & 77.3$^{\uparrow\textcolor{blue}{\textbf{2.3}}}$ & walk toward someone & 25.0 & 28.1$^{\uparrow\textcolor{blue}{\textbf{3.1}}}$ \\
 & fishing & 75.0 & 75.0 & open an umbrella & 28.6 & 42.9$^{\uparrow\textcolor{blue}{\textbf{14.3}}}$ \\ 
UAV-Human & put hands on hips & 73.3 & 73.3 & close an umbrella & 30.0 & 45.0$^{\uparrow\textcolor{blue}{\textbf{15.0}}}$ \\  
CSv1 & sit down & 72.7 & 72.7 & walk & 31.8 & 36.4$^{\uparrow\textcolor{blue}{\textbf{4.6}}}$ \\ 
 & take off a hat & 68.9 & 68.9 & give something to someone & 35.5 & 35.5 \\  
 & stand up & 68.2 & 79.5$^{\uparrow\textcolor{blue}{\textbf{11.3}}}$ & stab someone with a knife & 35.5 & 48.4$^{\uparrow\textcolor{blue}{\textbf{12.9}}}$ \\ 
 & open a box & 66.7 & 66.7 & threat some with a knife & 36.4 & 39.4$^{\uparrow\textcolor{blue}{\textbf{3.0}}}$ \\ 
 & apply cream to face & 65.9 & 68.2$^{\uparrow\textcolor{blue}{\textbf{2.3}}}$ & run & 36.7 & 36.7 \\
\hline 
\end{tabular}
\label{tab:ms_top10}
\end{table*}

\begin{table}[t]
\caption{Accuracy of multi-stream ensemble when using different modality. In the NTU 60 and NTU 120 datasets, we report the Top-1 accuracy. The HP represents heatmap pooled features.}
\centering
\begin{tabular}{ccccc}
\hline
\multirow{2}{*}{\textbf{Modality}} & \multicolumn{2}{c}{\textbf{NTU 60} ($\%$)} & \multicolumn{2}{c}{\textbf{NTU 120} ($\%$)} \\
\cmidrule{2-3}\cmidrule{4-5}
& {\textbf{X-Sub}}  & \textbf{X-View} & \textbf{X-Sub} & \textbf{X-Set} \\\hline
$l_1$: HP$_J$ & 93.8 & 94.5 & 86.1 & 88.8 \\
$l_2$: HP$_B$ & 92.8 & 92.8 & 85.3 & 87.4 \\
$l_3$: HP$_{JM}$ & 91.5 & 93.3 & 83.2 & 85.7 \\
$l_4$: HP$_{BM}$ & 90.7 & 92.8 & 82.0 & 84.3 \\
$l_5$: HP+RGB+Text & 97.0 & 99.0 & 96.3 & 96.5 \\\hline
$l1$+$l5$ & 97.3 & 99.2 & 96.4 & 96.8 \\
$l2$+$l5$ & 97.1 & 99.1 & 96.4 & 96.7 \\
$l1$+$l2$ & 94.3 & 95.3 & 87.8 & 90.2 \\
$l1$+$l2$+$l5$ & 97.5 & 99.4 & 96.5 & 97.0 \\
$l1$+$l2$+$l3$+$l4$+$l5$ & \textbf{97.9} & \textbf{99.7} & \textbf{96.8} & \textbf{97.5} \\
\hline
\end{tabular}
\label{tab:Ensemble_NTU}
\end{table}

\begin{table}[t]
\caption{Accuracy of multi-stream ensemble when using different modality. In the UAV-Human dataset, we report the Top-1 accuracy, while in the Toyota-Smarthome dataset, we report the mean per-class accuracy. The HP represents heatmap pooled features.}
\centering
\begin{tabular}{cccccc}
\hline
\multirow{2}{*}{\textbf{Modality}} & \multicolumn{3}{c}{\textbf{Toyota-Smarthome} ($\%$)} & \multicolumn{2}{c}{\textbf{UAV-Human} ($\%$)}\\
\cmidrule{2-4} \cmidrule{5-6}
& \textbf{CS} & \textbf{CV$_{1}$} & \textbf{CV$_{2}$} & \textbf{CSv1} & \textbf{CSv2} \\\hline
$l_1$: HP$_J$ & 51.5 & 30.9 & 37.0 & 33.7 & 56.5 \\
$l_2$: HP$_B$ & 51.3 & 29.9 & 30.3 & 31.3 & 52.6 \\
$l_3$: HP$_{JM}$ & 45.2 & 29.5 & 34.5 & 28.2 & 49.3 \\
$l_4$: HP$_{BM}$ & 43.2 & 28.6 & 33.9 & 25.5 & 44.3 \\
$l_5$: HP+RGB+Text & 81.3 & 51.7 & 66.4 & 52.0 & 77.8 \\\hline
$l1$+$l5$ & 81.4 & 53.0 & 66.5 & 53.2 & 79.2 \\
$l2$+$l5$ & 81.1 & 53.9 & 66.3 & 52.9 & 78.8 \\
$l1$+$l2$ & 55.7 & 32.1 & 37.1 & 36.6 & 60.5 \\
$l1$+$l2$+$l5$ & 81.7 & 53.6 & 66.8 & 53.4 & 79.5 \\
$l1$+$l2$+$l3$+$l4$+$l5$ & \textbf{81.6} & \textbf{57.9} & \textbf{67.4} & \textbf{53.6} & \textbf{80.2} \\
\hline
\end{tabular}
\label{tab:Ensemble_SH}
\end{table}

\begin{figure}[t]
  \centering
   \includegraphics[width=1.0\linewidth]{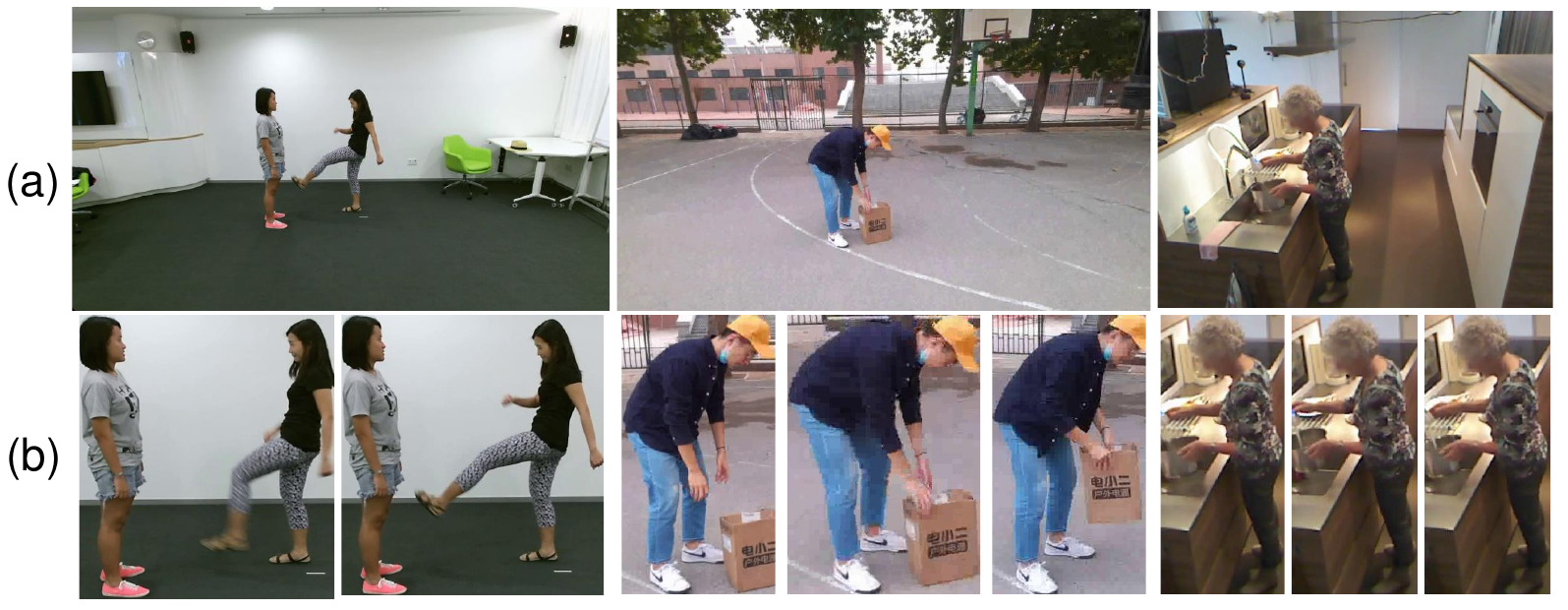}
   \caption{Visualization of different action samples with and without using a human detector. (a) We present the RGB images of datasets NTU-RGB+D 120, UAV-Human, and Toyota-Smarthome without using a human detector. (b) We present the RGB images of datasets NTU-RGB+D 120, UAV-Human, and Toyota-Smarthome with a human detector filtering environmental noise.}
   \label{fig:detector}
\end{figure}

\begin{figure}[t]
  \centering
   \includegraphics[width=1.0\linewidth]{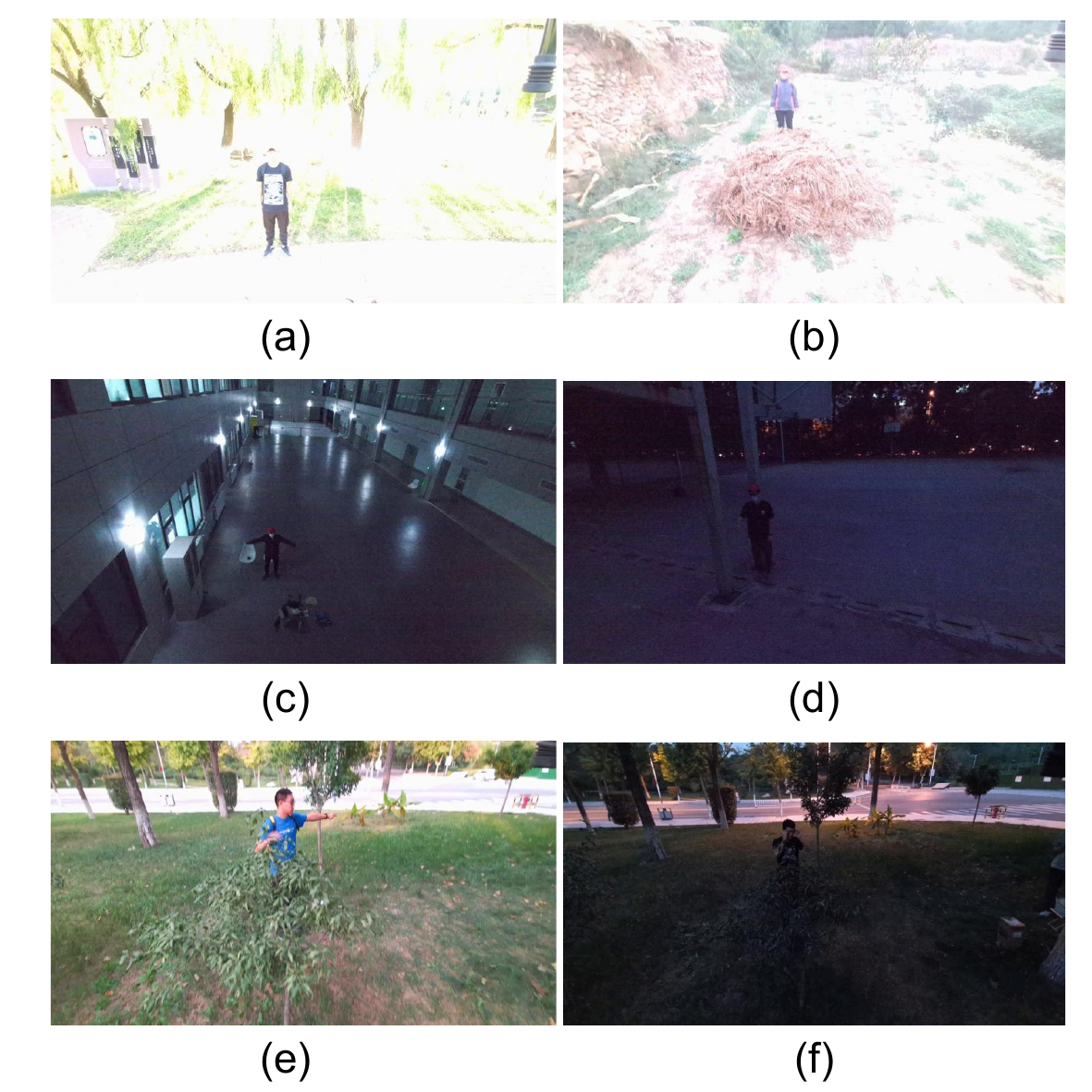}
   \caption{Visualization of some action samples under extreme scenarios. The action samples in subfigures (a) and (b) are captured under overexposed scenarios. The action samples in subfigures (c) and (d) are captured under low-light night scenes. The action samples in subfigures (e) and (f) are captured under occlusion scenes. }
   \label{fig:uav_extreme}
\end{figure}

\section{Experiments}
\subsection{Datasets}
We conduct experiments on four well-established large-scale human action recognition datasets: \textbf{NTU-RGB+D 60} \cite{7780484}, \textbf{NTU-RGB+D 120} \cite{8713892}, \textbf{Toyota-Smarthome} \cite{das2019toyota, dai2022toyota} and \textbf{UAV-Human} \cite{Li_2021_CVPR}. The following sections provide a detailed introduction to these datasets and their evaluation strategies.

\textbf{NTU-RGB+D 60} dataset \cite{7780484} (NTU 60) is a prominent benchmark for action recognition, featuring 56,880 skeleton/pose sequences and videos of human actions. These actions are grouped into 60 distinct categories and performed by 40 unique individuals. The original paper introduces two evaluation protocols: (1) \textbf{Cross-view} (X-View), where training data is collected from cameras positioned at 0° (view 2) and 45° (view 3) and testing data comes from a -45° (view 1) perspective. (2) \textbf{Cross-subject} (X-Sub), where 20 subjects' data is used for training while the other 20 subjects are reserved for testing.

\textbf{NTU-RGB+D 120} dataset \cite{8713892} (NTU 120) is an extension of the NTU-RGB+D dataset, which consists of 114,480 video samples spanning 120 action classes and is performed by 106 participants. The evaluation follows two benchmarks introduced in the original paper \cite{8713892}: (1) \textbf{Cross-subject} (X-Sub), where data from 53 participants is used for training and the remaining 53 for testing. (2) \textbf{Cross-setup} (X-Set), where training samples are drawn from even-numbered setups, while odd-numbered setups are reserved for testing.

\textbf{Toyota-Smarthome} dataset \cite{das2019toyota, dai2022toyota} is a video dataset recorded in an apartment equipped with Kinect cameras. It contains 31 daily living activities and 18 subjects. The videos were clipped per activity, resulting in a total of 16,115 video samples. We follow the protocol provided in the original paper \cite{das2019toyota} and use the same mean per-class accuracy for evaluation: \textbf{Cross-subject} (CS), where the training group consists of 11 subjects with IDs (\ie 3, 4, 6, 7, 9, 12, 13, 15, 17, 19, 25) and the remaining 7 subjects are reserved for testing. \textbf{Cross-view} (CV), is divided into $CV_{1}$ and $CV_{2}$, where the camera 2 is used for testing and camera 5 for validation. The $CV_{1}$ selects all samples from camera 1 for training, while the $CV_{2}$ includes samples from multiple cameras: 1, 3, 4, 6 and 7 for the training set.

\textbf{UAV-Human} dataset \cite{Li_2021_CVPR} is a comprehensive dataset for action recognition, consisting of 22,476 video clips spanning 155 unique classes. Captured by a UAV navigating diverse urban and rural environments under both day and night conditions, the dataset provides a vivid portrayal of human activities. It features 119 individuals performing 155 different actions across 45 distinct settings, offering a rich and diverse representation of human behavior. For evaluation, the original study proposes two benchmark scenarios (CSv1 and CSv2), with data from 89 participants used for training and 30 for testing, ensuring robust and reliable analysis.

\subsection{Implementation Details}
All experiments are conducted on four NVIDIA GeForce RTX 3080 Ti GPUs. Our proposed Heatmap Pooling Network (HP-Net) uses YoloV5 \cite{yolov5} as the human detector for videos and utilizes SimCC \cite{li2022simcc} as the human pose estimator. In detail, we initialize the SimCC model with the pretrained weights of an HR-Net backbone trained on the COCO dataset, using an input resolution of $256\times192$. Besides, our HP-Net uses the visual encoder provided in XCLIP \cite{XCLIP} and the text encoder from CLIP \cite{radford2021learning}, initializing both with pre-trained parameters on Kinetics-600 \cite{kay2017kinetics} as provided in \cite{XCLIP}. To effectively model the heatmap pooled features, our HP-Net selected the lightweight CTR-GCN \cite{Chen_2021_ICCV} as the topology model backbone. We use weight decay with adam (AdamW) to train our model for a total number of 30 epochs with batch size 4, while keeping the text encoder fixed. Inspired by \cite{MMNet, PoseC3D}, in addition to using the single-stream HP-Net shown in \figref{fig:figure2}, we also employ CTR-GCN \cite{Chen_2021_ICCV} to perform multi-stream modeling on the extracted heatmap pooled features. We retain classification scores from all streams and perform the same late fusion strategy \cite{MMNet, PoseC3D, EPPNet} to achieve multi-stream ensemble. More detailed implementation can be found in our open-source code project.

\subsection{Comparison with State-of-the-Art Methods}
In Table \ref{tab:sota} and Table \ref{tab:uav}, we compare our Heatmap Pooling Network (HP-Net) with the state-of-the-art methods on the NTU RGB+D 60, NTU RGB+D 120, Toyota-Smarthome and UAV-Human datasets. On four datasets, our model outperforms all existing methods under nearly all evaluation benchmarks. Our HP-Net using only a single stream, achieves recognition accuracies of 96.3$\%$ and 96.5$\%$ on the X-Sub and X-Set benchmarks of the NTU RGB+D 120 dataset, which outperforms the $\pi$-ViT \cite{reilly2024pivit} by 1.2$\%$ and 0.4$\%$ respectively. When adopting the same late-fusion strategy for multi-stream ensemble, our HP-Net achieves accuracies of 96.8$\%$ and 97.5$\%$ on the aforementioned benchmarks, which outperforms the PoseConv3D \cite{PoseC3D} by 1.5$\%$ and 1.1$\%$ respectively. Notably, unlike PoseC3D \cite{PoseC3D} that uses heavy CNNs to model 3D heatmaps, our HP-Net employs a lightweight GCN to model heatmap pooled features and achieves higher recognition accuracy when fused with RGB data.

In addition, our HP-Net achieves recognition accuracies of 97.9$\%$ and 99.7$\%$ on the X-Sub and X-View benchmarks of the NTU RGB+D 60 dataset, which outperforms the DVANet by 4.5$\%$ and 1.5$\%$ respectively. Compared to MMNet \cite{MMNet} that performs late fusion of 3D pose and RGB video, our HP-Net surpasses it by 1.9$\%$ and 0.9$\%$ on the above benchmarks respectively. On the Toyota-Smarthome dataset, our HP-Net also achieved state-of-the-art action recognition performance, with mean per-class accuracy of 81.6$\%$, 57.9$\%$ and 67.4$\%$ across three benchmarks. In the UAV-Human dataset, our HP-Net achieves accuracy rates of 53.6$\%$ and 80.2$\%$ on two benchmarks, outperforming existing methods. The state-of-the-art recognition accuracy across the four datasets above fully demonstrates the effectiveness and superiority of our proposed HP-Net, as it outperforms existing methods.

\subsection{Ablation Studies}
In this section, we conduct ablation experiments on the most challenging X-Sub benchmark from NTU-RGB+D 120 dataset to demonstrate the effectiveness, robustness and simplicity of the individual modules in our proposed Heatmap Pooling Network (HP-Net). Firstly, we thoroughly explored the effectiveness of the proposed Feedback Pooling Module (FPM). Then, we investigated the transferability of the heatmap pooled features generated based on the FPM. Finally, we examined the impact of the settings of some hyperparameters and modules in the HP-Net on the performance.
\subsubsection{Effectiveness of FPM}
In Table \ref{tab:FPM}, we model the outputs from the same pose estimation process using different visual backbones (\eg GCN and Transformer) to demonstrate the effectiveness and robustnes of our proposed feedback pooling module (FPM). The 2D Pose in Table \ref{tab:FPM} represents the human skeleton data directly obtained from videos using pose estimation, while the Pooled Feature refers to the concise features obtained by pooling heatmaps using 2D pose data based on our proposed FPM. Meanwhile, we convert the 2D pose and heatmap pooled features into four modalities (\ie Joint, Bone, Joint Motion and Bone Motion) to comprehensively demonstrate the effectiveness of the heatmap pooled features extracted by the FPM.

In addition, we use two classic network backbones (\eg CTR-GCN \cite{Chen_2021_ICCV} and Skeleton-MixFormer \cite{xin2023skeleton} ) to model the heatmap pooled features, demonstrating that the pooled features extracted by our FPM can be easily transferred to different visual backbones. Unlike precious methods that require heavy CNNs to model heatmaps, the heatmap pooled features obtained by our FPM can be effectively modeled using lightweight GCNs and Transformers. The results in Table \ref{tab:FPM} show that the heatmap pooled features extracted using FPM outperform the directly generated 2D pose data in terms of recognition performance, across different modalities and network backbones. Compared to the conventional 2D poses estimated directly from heatmaps, our FPM improves recognition accuracy in the Joint modality by 3.1$\%$ and 3.0$\%$ when modeled with different network backbones.

To thoroughly demonstrate the effectiveness of FPM, we compare the extracted heatmap pooled features with 3D pose data collected from sensors and heatmap coordinate vectors obtained by pose estimation, which is shown in Table \ref{tab:modality}. To ensure a fair comparison, we use CTR-GCN \cite{Chen_2021_ICCV} as the consistent modeling backbone. The results in Table \ref{tab:modality} demonstrate that our heatmap pooled features also exhibit outstanding performance advantages over 3D pose data and heatmap coordinate vectors. The results in Tables \ref{tab:FPM} and \ref{tab:modality} fully validate the effectiveness of our FPM, as the heatmap pooled features it extracts outperform other common features obtained from videos in recognition performance, while requiring only lightweight downstream networks for modeling.

In Table \ref{tab:action_acc}, we further analyze the effectiveness of the proposed heatmap pooled features for some action categories in the NTU120 dataset, where the action labels start from 1. To ensure a fair comparison, we use the same backbone modeling network to compare the proposed heatmap pooled features with the 2D pose data obtained from the same pose estimation network. Additionally, we include a comparison with the 3D pose data collected by Kinect cameras provided by the original dataset. Among the 35 action categories in Table 4, our proposed heatmap pooled features significantly improve recognition accuracy compared to both 2D and 3D pose data. Notably, for challenging actions involving environmental context or objects, our proposed heatmap pooled features demonstrate more significant performance advantages. For instance, in the action category \textit{play with phone/tablet} (action label is 29), the accuracy of heatmap pooled features surpasses that of 2D and 3D poses by 20.4$\%$ and 12.4$\%$, respectively. Similar examples include fine-grained action categories such as \textit{cutting nails} (action label is 75) and \textit{blow nose} (action label is 105).

In Table \ref{tab:HPE_Acc}, we explored the performance of heatmap pooled features obtained using different human pose estimation models. In terms of implementation, we first selected two pose estimation models, HR-Net \cite{wang2020deep} and SimCC \cite{li2022simcc}, to extract heatmap pooled features from the videos. Then, we utilized CTR-GCN \cite{Chen_2021_ICCV}, Skeleton-MixFormer \cite{xin2023skeleton} and the proposed HP-Net to model the extracted heatmap pooled features and compared their recognition performance. The result in Table \ref{tab:HPE_Acc} indicates that the heatmap pooled features extracted using SimCC demonstrate better performance, with recognition accuracy higher than HR-Net by 0.8$\%$ and 1.9$\%$ under the same network modeling. Therefore, in our HP-Net, we selected the SimCC model as the human pose estimation network. It is worth emphasizing that our HP-Net is orthogonal to the pose estimation model, allowing it to be easily transferred to different heatmap-based human pose estimation networks.

In Table \ref{tab:scale}, we also explore the recognition performance when using the FPM to pool heatmaps of different scales and modeling them with the same network backbone. The second column in Table \ref{tab:scale} represents the size of the heatmaps, while the third column indicates the size of the heatmap pooled features, where $n$ denotes the number of joints used for feedback pooling module (FPM). We found that the heatmap pooled features derived from human heatmap $H_2$ achieved better recognition performance, with an accuracy of  86.2$\%$ on the NTU-120 X-Sub benchmark. Therefore, in our proposed FP-Net, we exclusively use heatmap $H_2$ to derive the heatmap pooled features. In the specific pose estimation network we used, the size of heatmap $H_2$ is $[c_1=96, h_1=32, w_1=24]$, and the resulting heatmap feature's size is $[n=17, c_1=96]$. In Table \ref{tab:scale}, we compare the recognition performance of the three-scale heatmaps ($H_1$, $H_2$ and $H_3$), where the resolution decreases while the channel dimensionality increases progressively. In image modeling, high-resolution but low-dimensional representations tend to preserve richer spatial details (\eg, edges, contours, and textures as shown in Fig. \ref{fig:heatmap}), which primarily correspond to low-level visual features. In contrast, low-resolution but high-dimensional representations capture more abstract semantic information while losing a considerable amount of spatial detail. Heatmap $H_2$ strikes a balance between resolution and channel dimensionality, preserving sufficient spatial details while simultaneously extracting richer semantic information, which explains its superior recognition performance. To further validate the suitability of $H_2$ as the heatmap for extracting pooled features, we conducted a quantitative analysis based on the Fisher criterion \cite{gu2012generalized}. Specifically, we computed the Fisher scores of $H_1$, $H_2$ and $H_3$ on the NTU-RGB+D dataset, where the Fisher score is derived from intra-class and inter-class scatter, and serves as a widely adopted metric for evaluating feature discriminability in classification tasks. As illustrated in Fig. \ref{fig:fisher}, heatmap $H_2$ achieves the highest Fisher score, indicating stronger discriminability and greater suitability for tasks such as action recognition. This finding is consistent with the experimental results reported in Table \ref{tab:scale}.

In Table \ref{tab:stage}, we also explore the recognition performance when using the FPM to pool heatmaps of different stages in the human pose estimation process. In Table \ref{tab:region}, we explore the recognition performance when performing pooling over heatmaps using regions of varying sizes. Additionally, we analyze the parameters and computational cost required for modeling different forms of heatmap features in Table \ref{tab:param}. Unlike traditional methods \cite{Liu_2018_CVPR, PoseC3D} that convert heatmap features into image format and rely on heavy CNNs for modeling, our heatmap pooled features can be efficiently modeled using a lightweight GCN, demonstrating superior performance in both parameters and computational cost. In Table \ref{tab:pose-guided}, we investigate the performance of using pose-guided regional sampling on heatmaps and RGB images. The results in Table \ref{tab:pose-guided} demonstrate that our proposed feedback pooling mechanism for heatmap aggregation achieves superior performance compared to directly applying regional sampling on RGB images.

\subsubsection{Transferability of FPM}
In our heatmap pooling network (HP-Net), we designed two modules (\ie TRMM and SMCLM) to effectively fuse the heatmap pooled features generated by the FPM with the text and RGB modalities. In Table \ref{tab:TRMM+SMCLM}, we conducted ablation experiments on the designed TRMM and SMCLM, where the baseline retains only the text encoder and visual encoder. Integrating the heatmap pooled features generated by FPM via TRMM improved recognition accuracy by 0.4$\%$, while the addition of the SMCLM further boosted accuracy by 1.0$\%$. Additionally, we fused the heatmap pooled features generated by the FPM with text features using Cross Attention \cite{10.5555/3295222.3295349} and Feature-wise Linear Modulation (FiLM) \cite{perez2018film}, achieving improvements in recognition performance, which demonstrates the versatility and transferability of the heatmap pooled features. The results in Table \ref{tab:TRMM+SMCLM} fully validate the effectiveness of the text refinement modulation module (TRMM) and spatial-motion co-learning module (SMCLM). By effectively integrating with text and RGB modalities, our HP-Net achieved more accurate action recognition and demonstrated the transferability of the heatmap pooled features.

\subsubsection{Setting of HP-Net}
In our Heatmap Pooling Network (HP-Net), we use a human detector to extract human frames from RGB videos, aiming to filter out irrelevant environmental noise and focus on modeling human body parts. In Table \ref{tab:Detection}, we present the performance differences with and without using the human detector on the NTU 120 and Smarthome datasets. On the X-Sub benchmark of the NTU120 dataset, our HP-Net achieved a 9.3$\%$ increase in recognition accuracy when using a human detector compared to not using one. We believe that the performance difference is due to the large pixel space occupied by environmental factors in the original NTU120 dataset, which leads the visual encoder to model too much irrelevant environmental information. When a human detector is used to filter out most of the environmental factors, the visual encoder can focus more on modeling the pixel space where the human body is located, thereby improving the accuracy of human action recognition. Meanwhile, in \figref{fig:detector}, we show RGB images of different action samples with and without using the human detector. Besides, in Table \ref{tab:lambda}, we also explored the impact of different hyperparameters in the loss function on model performance.

\subsubsection{Robustness of HP-Net}
The UAV-Human dataset contains numerous action samples captured under extreme conditions, such as challenging weather, low-light night scenes, overexposure, and occlusion, making it highly challenging. In \figref{fig:uav_extreme}, we present several action samples captured under extreme scenarios. To further investigate the robustness of our HP-Net in recognizing actions under these extreme conditions, we conducted experiments on these samples using HP-Net, as reported in Table \ref{tab:uav_extreme}. When only standard RGB and text features were employed, the recognition accuracy was 42.3$\%$. By incorporating our HP-Net with heatmap pooled features, the accuracy increased to 51.5$\%$, resulting in an overall improvement of 9.2$\%$. These results in Table \ref{tab:uav_extreme} demonstrate that our HP-Net, along with the extracted heatmap pooled features, achieves good performance gains even under extreme conditions, thereby validating the effectiveness and robustness of the proposed framework.

\subsection{Multi-stream ensemble}
Multi-stream ensemble is a commonly used technique in human action recognition \cite{lee2022hierarchically, myung2024degcn, MMNet, Chen_2021_ICCV, xin2023skeleton, 10113233, PoseC3D}, which achieves better performance by integrating classification scores from multiple modalities through a late-fusion strategy. In Table \ref{tab:Ensemble_NTU} and Table \ref{tab:Ensemble_SH}, we present the recognition accuracy of multi-modality fusion using multi-stream ensemble on the NTU 60, NTU 120, Smarthome and UAV-Human datasets. In the implementation, we follow the approach in \cite{10113233} to convert the heatmap pooled features into four modalities: HP$_J$, HP$_B$, HP$_{JM}$ and HP$_{BM}$ as shown in the Tables \ref{tab:Ensemble_NTU} and \ref{tab:Ensemble_SH}. We then model each of these modalities separately using the same network (i.e. CTR-GCN \cite{Chen_2021_ICCV}), preserving their classification scores. Finally, we use a late-fusion strategy \cite{das2020vpn,das2021vpn++, MMNet, PoseC3D} to combine the classification scores from the four modalities with the classification scores obtained from the HP-Net in \figref{fig:figure2}, aiming to achieve better recognition performance. The experimental results in Tables \ref{tab:Ensemble_NTU} and \ref{tab:Ensemble_SH} show that the heatmap pooled features we proposed also exhibit strong transferability. By using a late fusion strategy for multi-stream ensemble, they enable more robust action recognition. In Table \ref{tab:ms_top10}, we have displayed the top-10 action samples with the highest and lowest accuracy in the single-stream HP-Net (\ie HP-Net (1s)). Meanwhile, we have shown the improvement in the recognition accuracy of these action samples after using multi-stream ensemble (\ie HP-Net (ms)). The results in Table \ref{tab:ms_top10} indicate that the accuracy of both simple and difficult actions has improved after using multi-stream ensemble, which also demonstrates that multi-stream ensemble based on our proposed heatmap pooled features will achieve better performance.

\subsection{Visualization}
In Table \ref{tab:scale}, we compare the recognition performance of heatmap pooled features obtained from human heatmaps of different scales. Meanwhile, as shown in Fig. \ref{fig:heatmap}, we present the visualizations of the above three-scale heatmaps from Table \ref{tab:scale} for various action samples, corresponding to the second, third and fourth columns of each subplot, respectively. Specifically, we selected 12 action samples from the NTU60 (\eg (a) \textit{drink water}, (b) \textit{jump
up}, (c) \textit{hand waving}, (d) \textit{clapping}), NTU120 (\eg (e) \textit{shoot at basket}, (f) \textit{squat down}, (g) \textit{butt kicks}, (h) \textit{put on bag}), Toyota-Smarthome ((i) \textit{Cook.Cleandishes}, (j) \textit{Usetelephone}) and UAV-Human ((k) \textit{cough}, (l) \textit{move a box}) datasets for visualization. In each subplot, the first column represents the original RGB image and the 2D pose data obtained through pose estimation, while the fifth column illustrates the heatmap pooled features proposed in our Heatmap Pooling Network (HP-Net).

In heatmap-based pose estimation, models typically extract 2D pose data from videos, which is then used for human action recognition. Unlike previous methods that directly rely on 2D pose data, our HP-Net innovatively incorporates a feedback pooling mechanism to extract compact, informative and robust heatmap pooled features from the pose estimation process. These extracted features offer superior robustness and accuracy in action recognition, outperforming traditional 2D pose representations. In Fig. \ref{fig:acc}, we use the same network backbone to model both the 2D pose and the heatmap pooled features derived from the same pose estimation model. We then visualize the top-8 action categories on benchmarks NTU 120 XSub, Toyota-Smarthome CS and UAV-Human CSv1 where the heatmap pooled features outperform the 2D pose in recognition accuracy. We also visualize the overall recognition accuracy of the two modalities across all action categories in Fig. \ref{fig:acc}.

In Fig. \ref{fig:tsne}, we showcase the differences between 2D/3D pose data and heatmap pooled features from a feature perspective. Specifically, we use the same network to model the three types of data, where 2D pose data and heatmap pooled features come from the same pose estimation network, while 3D pose data is provided by the original dataset. In detail implementation, we selected samples from 60 actions in the NTU60 X-Sub benchmark for T‐distributed stochastic neighbor embedding (T-SNE) visualization. On the premise of ensuring fairness, we found that the modeled heatmap pooled features exhibit better feature discriminability. The sample features from different action categories are more dispersed and distinguishable in the feature space, which also demonstrates the superiority of our proposed heatmap pooled features.

\section{Discussion}
In this paper, we propose a Heatmap Pooling Network (HP-Net), whose core motivation is to leverage a feedback mechanism to extract heatmap pooled features that can effectively represent human motion from videos. Extensive experiments conducted in controlled laboratory settings, home environments, and aerial surveillance scenarios comprehensively validate the effectiveness, robustness, and transferability of the heatmap pooled features. It is worth emphasizing that the core idea of HP-Net is not limited to the three aforementioned scenario-specific datasets, but rather represents a general approach to human motion analysis. For example, HP-Net can also be applied to generic action recognition benchmarks such as Something-Something \cite{goyal2017something} to enhance action recognition performance. Below, we discuss the potential, adaptability and challenges of HP-Net on general action recognition benchmarks.

First, the feedback pooling mechanism employed by HP-Net does not rely on datasets from specific scenarios. Whenever heatmap-based pose or gesture estimation methods are used to extract video features for human motion analysis, the feedback pooling mechanism of HP-Net can be applied. For instance, the Something-Something dataset is an action dataset that involves interactions between human hands and objects. By employing a heatmap-based gesture estimation model to obtain gesture poses and gesture heatmaps, the estimated gesture poses can be used to feedback-pool the gesture heatmaps, thereby generating gesture-specific heatmap pooled features that help improve recognition performance. Second, the heatmap pooled features extracted by HP-Net are influenced by different heatmap-based pose estimation models. This conclusion that has been thoroughly demonstrated in the ablation experiments presented above (\ie in Table \ref{tab:HPE_Acc}). Therefore, when the heatmap-based pose estimation model fails to accurately estimate poses and heatmaps from a general action dataset, the extracted heatmap pooled features may not provide significant performance improvements for the corresponding tasks.  However, this potential risk does not constrain the feedback pooling concept proposed by the HP-Net, as the aforementioned issue can be effectively addressed by employing more advanced heatmap-based pose estimation networks.

In conclusion, the feedback pooling mechanism of HP-Net presents a promising and generalizable solution for action recognition from RGB videos. It can efficiently extract compact and robust heatmap pooled features, making it a strong candidate for human motion analysis. Exploring the applicability of the feedback pooling mechanism across different human motion analysis tasks represents a clear and valuable direction for our future work, and we are optimistic about the results based on its principled design.

\section{Conclusion}
We present a novel Heatmap Pooling Network (HP-Net) for action recognition from RGB videos, which extracts information-rich, robust and concise heatmap pooled features through a Feedback Pooling Module (FPM). The innovative FPM is compatible with various human pose estimation methods, forging a new link between human pose estimation and video action recognition. The HP-Net extends image-based heatmap features to support a wider range of visual backbone networks while enabling seamless integration with other multimodal features for more robust and precise human action recognition. Compared to traditional modality data, the heatmap pooled features in the HP-Net establish a new modality benchmark for video action recognition. The effectiveness of our proposed HP-Net is verified on four benchmark datasets namely NTU RGB+D 60, NTU RGB+D 120, Toyota-Smarthome and UAV-Human, where it outperforms state-of-the-art methods.  

\bibliographystyle{IEEEtran}
\bibliography{HP-Net}

\begin{IEEEbiography}[{\includegraphics[width=1in,height=1.25in,clip,keepaspectratio]{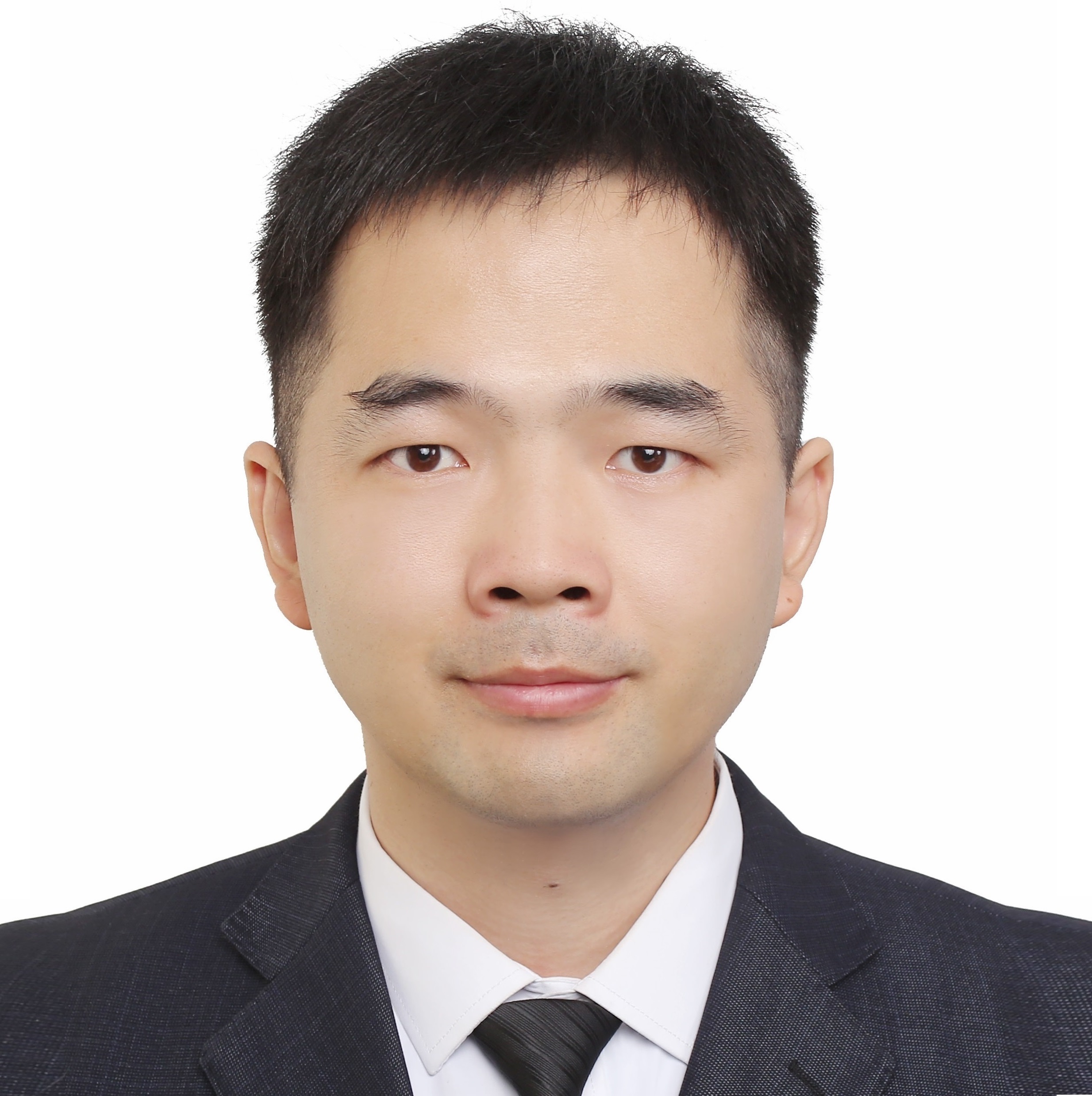}}]{Mengyuan Liu}
received his Ph.D. degree from the School of Electrical Engineering and Computer Science, Peking University. He was a research fellow at the School of Electrical and Electronic Engineering, Nanyang Technological University. Currently, he is an Assistant Professor at Peking University, Shenzhen Graduate School. His research focuses on human-centric perception and robot learning. His work has been published in leading conferences and journals, including CVPR and T-PAMI. He also serves as an Associate Editor for PR.
\end{IEEEbiography}

\begin{IEEEbiography}[{\includegraphics[width=1in,height=1.25in,clip,keepaspectratio]{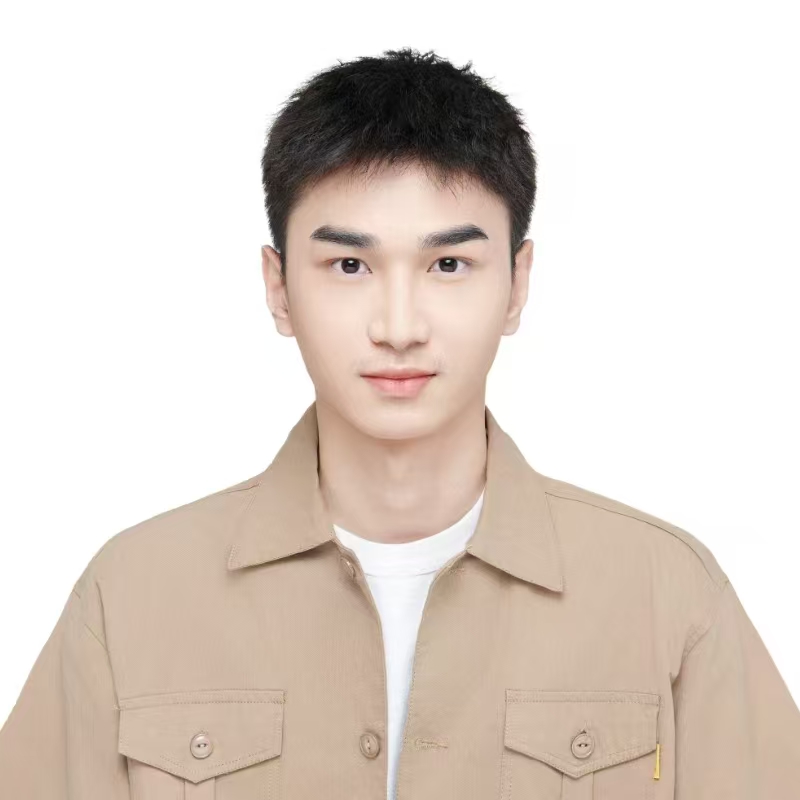}}]{Jinfu Liu}
received the M.Sc. degree from Sun Yat-sen University, Shenzhen, China, in 2025. He is currently an algorithm engineer with DJI Technology Co., Ltd. His work have been published in T-MM and ACM MM. His research interests include video understanding, content generation and on-device AI. He also serves as a reviewer for T-CSVT and ICME.
\end{IEEEbiography}

\begin{IEEEbiography}[{\includegraphics[width=1in,height=1.25in,clip,keepaspectratio]{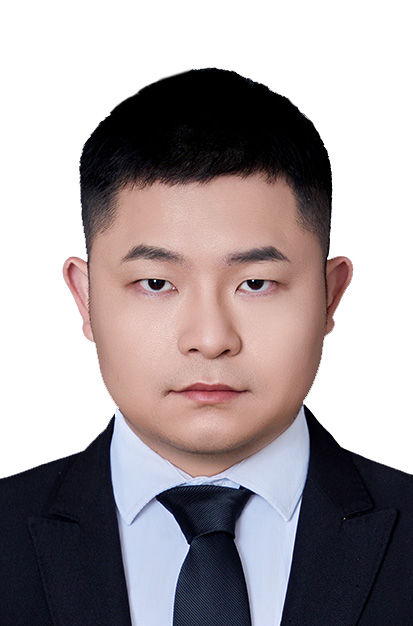}}]{Yongkang Jiang}
 received the B.Sc. degree and Ph.D. degree in mechanical and electronic engineering from Beihang University, Beijing, China, in 2015 and 2021, respectively. He is currently an assistant professor with the College of Electronics and Information Engineering, Tongji University, Shanghai, China. His research interests include computer vision and physical interaction.
\end{IEEEbiography}

\begin{IEEEbiography}[{\includegraphics[width=1in,height=1.25in,clip,keepaspectratio]{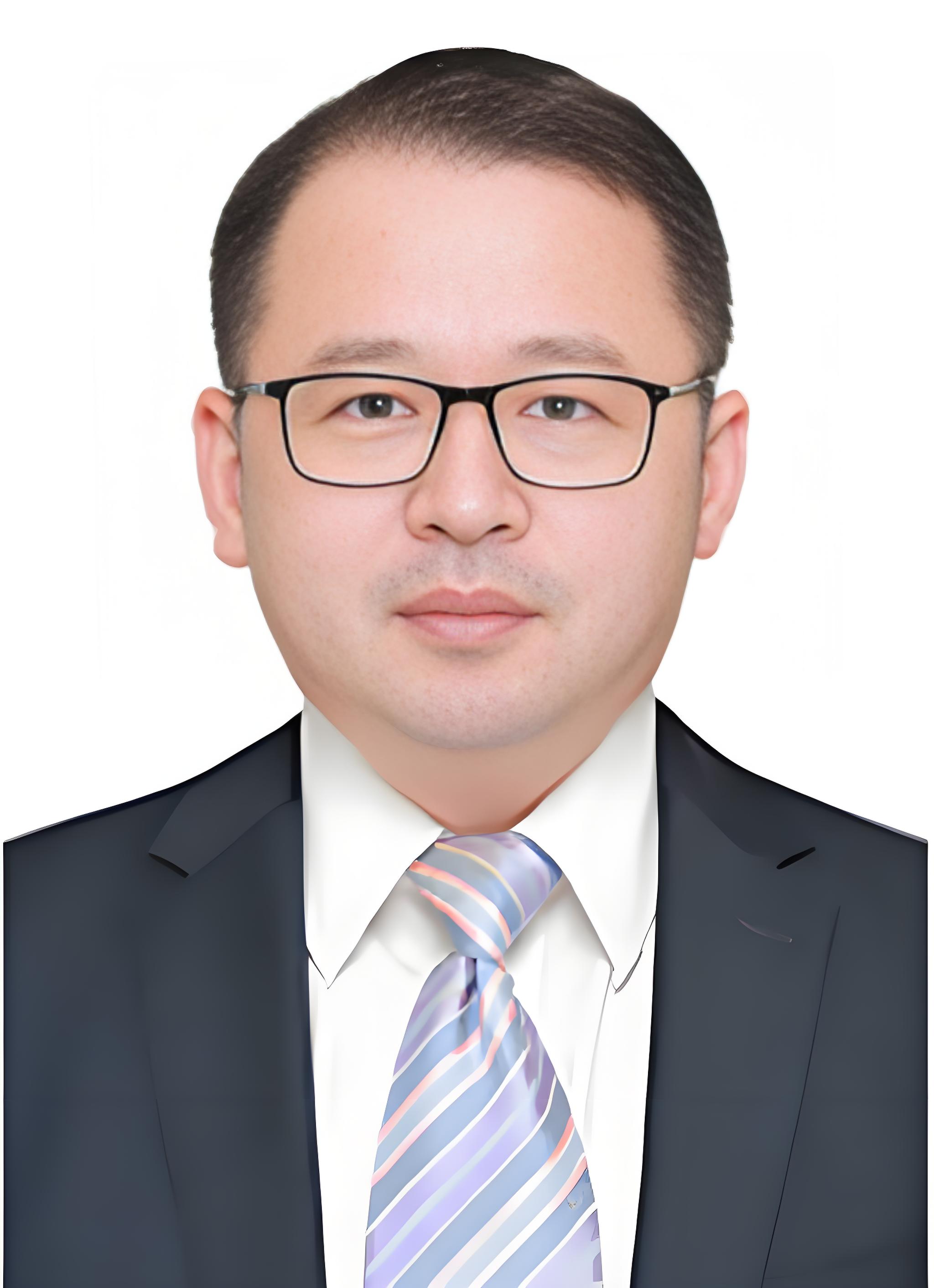}}]{Bin He}
 received the Ph.D. degree in mechanical and electronic control engineering from Zhejiang University, Hangzhou, China, in 2001. He held his post-doctoral research appointments with The State Key Laboratory of Fluid Power Transmission and Control, Zhejiang University, from 2001 to 2003. He is currently a Professor with the College of Electronics and Information Engineering, Tongji University, Shanghai, China. His current research interests include computer vision and intelligent robot.
\end{IEEEbiography}

\end{document}